\documentclass[letterpaper, 10 pt, conference]{ieeeconf}  
\usepackage{graphicx}
\usepackage{multirow}
\usepackage{amsmath}
\usepackage{subfigure}
\usepackage{setspace}
\usepackage{tabularx}
\usepackage{stackengine}
\usepackage{color}
\usepackage{cite}	
\newcolumntype{Y}{>{\centering\arraybackslash}X}
\IEEEoverridecommandlockouts
\title{\LARGE \bf
Parse Geometry from a Line: Monocular Depth Estimation with Partial Laser Observation
}

\author{Yiyi Liao$^{1}$, Lichao Huang$^{2}$,  Yue Wang$^{1}$, Sarath Kodagoda$^{3}$, Yinan Yu$^{2}$, Yong Liu$^{1}$ 
\thanks{$^{1}$Yiyi Liao, Yue Wang and Yong Liu are with the State Key Laboratory of Industrial Control Technology and Institute of Cyber-Systems and Control, Zhejiang University, Zhejiang, 310027, China. Yong Liu is the corresponding author of this paper, e-mail: \small  yongliu@iipc.zju.edu.cn.}
\thanks{$^{2}$Lichao Huang and Yinan Yu are with the Horizon Robotics, China.}
\thanks{$^{3}$Sarath Kodagoda is with the Centre for Autonomous Systems (CAS), The University of Technology, Sydney, Australia.}}

\begin{document}

\maketitle
\thispagestyle{empty}
\pagestyle{empty}

\begin{abstract}

Many standard robotic platforms are equipped with at least a fixed 2D laser range finder and a monocular camera. Although those platforms do not have sensors for 3D depth sensing capability,  knowledge of depth is an essential part in many robotics activities. Therefore, recently, there is an increasing interest in depth estimation using monocular images. As this task is inherently ambiguous, the data-driven estimated depth might be unreliable in robotics applications. In this paper, we have attempted to improve the precision of monocular depth estimation by introducing 2D planar observation from the remaining laser range finder without extra cost. Specifically, we construct a dense reference map from the sparse laser range data, redefining the depth estimation task as estimating the distance between the real and the reference depth. To solve the problem, we construct a novel residual of residual neural network, and tightly combine the classification and regression losses for continuous depth estimation. Experimental results suggest that our method achieves considerable promotion compared to the state-of-the-art methods on both NYUD2 and KITTI, validating the effectiveness of our method on leveraging the additional sensory information. We further demonstrate the potential usage of our method in obstacle avoidance where our methodology provides comprehensive depth information compared to the solution using monocular camera or 2D laser range finder alone.

\end{abstract}

\section{Introduction}

Depth information plays an important role in daily lives of human, and is also a valuable cue in computer vision and robotics tasks. Many research works have demonstrated the benefit of introducing the depth information for tasks such as object recognition and scene understanding~\cite{silberman2012indoor,liao2016understand,wang2016framework}. 
Recently, some researchers have opted to use monocular cameras to estimate the depths because of its inherent practical value. Monocular depth estimation is particularly challenging as it is a well known ill-posed problem. It is non trivial due to the vast amount of monocular depth cues, such as object sizes, texture gradients and overlaps needed for such depth estimations, in addition with the global scale of the scene.  Thanks to the development of the deep convolutional neural networks over the recent years, remarkable advances are achieved on the task of monocular depth estimation~\cite{eigen2014depth,eigen2015predicting,liu2015learning,laina2016deeper,cao2016estimating}. A possible reason is that the monocular cues can be better modeled with the larger capacity of the deep network. However, the global scale of the scene remains a major ambiguity in monocular depth estimation~\cite{eigen2014depth}. With this ambiguity, the depth estimation result might be unreliable for robotics applications such as obstacle avoidance.

\begin{figure}[tpb]
\centering
\includegraphics[height=4cm]{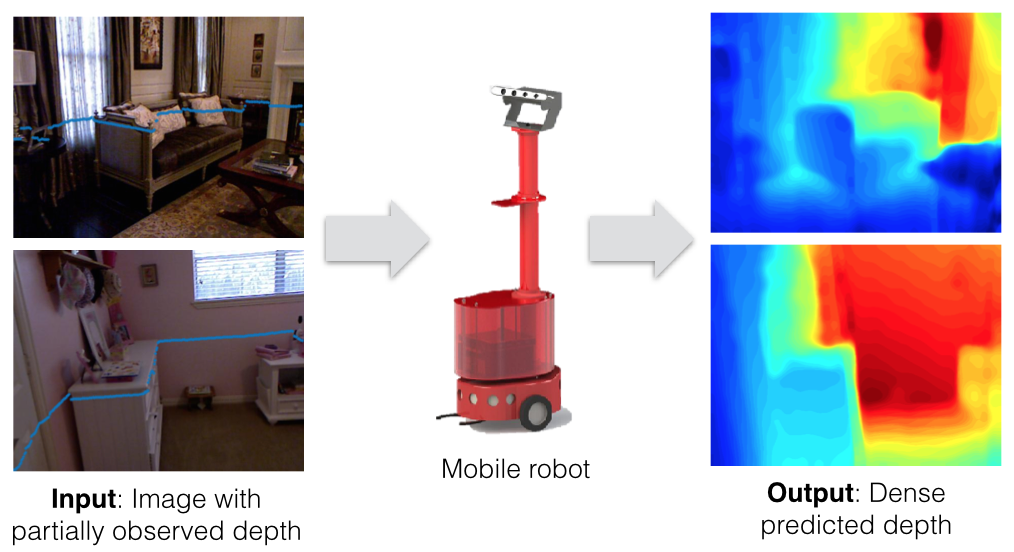}
\caption{ Illustration of our proposed method. The input of our method is a single image and a planar of 2D laser range data, obtained from a monocular camera and a 2D laser range finder. The aim of this paper is to precisely estimate the dense depth of the full scene. }
\label{fig_example}
\end{figure}

In this regard, a natural option to consider is to see whether the global ambiguity can be resolved using complementary sensory information. We want to exploit this idea by introducing limited direct depth observations to the monocular depth estimation task using a planar 2D laser range finder. With the availability of cheap 2D laser range finders, this option can be attractive in robotics applications in terms of accuracy and cost of sensing. Illustrative examples can be found in Figure~\ref{fig_example}. Ideally, the partially observed depth information can be employed to better estimate the global scale while the monocular image can be exploited for the relative depth estimation. 
To achieve this goal, we construct a novel convolutional neural network architecture to solve the partially observed depth estimation task. 

For mobile robots, the proposed configuration for partial depth observation is very common. Many well-known robots such as Pioneer\footnote{http://www.mobilerobots.com/ResearchRobots/PioneerP3DX.aspx} and K5 Security Robot\footnote{http://knightscope.com} are equipped with a camera and a 2D laser range finder. The 2D laser range finder is indispensable for navigation and obstacle avoidance on the mobile robot~\cite{cherubini2014autonomous,liaoplace}.
We demonstrate that our method facilitates greater perception for obstacle avoidance compared to that of a single 2D laser range finder, as the latter has a very limited vertical field of view which might be insufficient to completely reflect the surrounding environments especially with voids.
Therefore, our method is a painless extension for most mobile robots with no requirement of additional cost.

The key task of this paper is to effectively leverage the limited and sparse 2D laser data for precisely estimation of the completed and dense depth. We formulate this problem as an end-to-end learning task based on a novel fully convolutional neural network.
The contribution of this paper can be summarized as follows:
\begin{itemize}
\item We introduce the 2D laser range data to the task of monocular depth estimation by constructing a dense reference depth map from the sparse observation. With the dense reference map, the task of estimating the depth is redefined as estimating the residual depth between the real depth and the reference depth.
\item To explicitly estimate the residual depth, we construct a novel network architecture named residual of residual neural network.
Besides, the network combines both classification and regression losses for effectively estimating the continuous depth value. 
\item We conduct experiments on both indoor and outdoor environments and gain considerable promotion compared to the state-of-the-art monocular depth estimation methods, as well as another partially observed depth estimation method. We further demonstrate its potential usage in obstacle avoidance for mobile robots.
\end{itemize}
The remainder of the paper is organized as follow: Section~\ref{sec_related} gives a review of related works. The methodology for solving the partially observed depth estimation task is given in Section~\ref{sec_model}, and the experimental results are presented in Section~\ref{sec_exp}. Finally, we conclude the paper in Section~\ref{sec_conclu}.

\section{Related Works}\label{sec_related}

In recent years, deep learning methods are intensively exploited in depth estimation using single images~\cite{eigen2014depth,eigen2015predicting,liu2015learning,laina2016deeper,cao2016estimating}. Deep networks are validated to be more effective compared to the conventional methods based on hand-crafted features and graphical models~\cite{saxena2005learning,saxena2009make3d}. Eigen et al.~\cite{eigen2014depth} firstly proposed to regress the depth value in the end-to-end framework using deep neural network. That work was extended to simultaneously estimate the depths, surface normals and semantic labels in their latter work~\cite{eigen2015predicting}. Liu et al.~\cite{liu2015learning} proposed to combine Conditional Random Field (CRF) and the deep neural network for depth estimation. The deep neural network learned the unary and pairwise potentials and the CRF was jointly optimized with the network. 
More recently, Laina et al.~\cite{laina2016deeper} and Cao et al.~\cite{cao2016estimating} tackled the depth estimation task based on the Residual Neural Network (ResNet)~\cite{he2015deep}, which won the first place on the classification and detection competition tasks at ILSVRC and COCO in 2015. Specifically, Laina et al.~\cite{laina2016deeper} regressed the depth value using the fully convolutional ResNet, and a novel up-convolutional scheme was developed for fine-grained estimation. Instead of regression, Cao et al.~\cite{cao2016estimating} regarded the depth estimation as a classification task, where the estimation probability could be obtained for refinement using CRF. Taking the advantage of the deep ResNet, Laina et al.~\cite{laina2016deeper} and Cao et al.~\cite{cao2016estimating} set a new baseline in the depth estimation task. As can be seen, previous works indicate a trend of learning all the variations in depth estimation using one deeper model, which might be difficult for disambiguating the global scale. Differently, in this paper, we propose to take the advantage of an external but common sensor on mobile robots to get an relatively reliable estimation of the global scale, upon which the variation that need to be modeled by the network could be reduced. To solve this partially observed depth estimation task, we propose a novel architecture of fully convolutional network.

In robotics scenario, there are also attempts for depth estimation with partially observed depths. A popular topic is 
the dense depth reconstruction with fusion of sparse 3D laser range data and the monocular image~\cite{harrison2010image,pinies2015too}. As the 3D laser range data is obtained, this problem is usually formulated as an inpainting problem considering the measurement compatibility and the smoothness regularization. Both~\cite{harrison2010image} and~\cite{pinies2015too} made efforts on designing the regularization term and presented outstanding performances, where the former introduced a second order smoothness term and the latter searched optimal regularizer for different scenes. It is to be noted that the inpainting method requires the laser range data to cover most of the scene, which means it can only work with 3D laser range finder. With only a single planar view of 2D laser range data, the inpainting formulation is intractable due to severely insufficient information. In this paper, we state the partially observed task as a discriminative learning task, which can estimate the depth even with a planar view. A relatively similar work to this paper is~\cite{cadena2016multi}. The authors proposed to use a Multi-modal Auto-Encoder to impute the missing depth in the sparse depth map estimated by structure from motion. In this paper, we alternatively estimate the residual between the real depth and the reference depth, which is shown to be more effective in the experiments.




\begin{figure}[tpb]
\centering
\includegraphics[height=4.2cm]{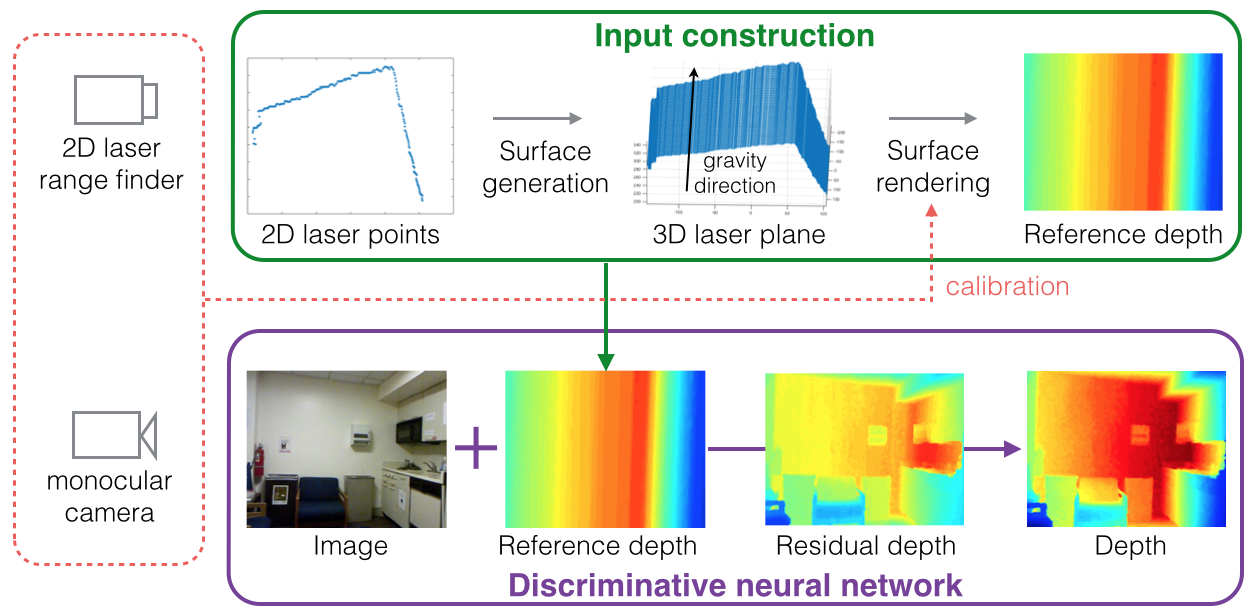}
\caption{Pipeline of the proposed method. A 3D surface is generated from the 2D laser scan along the gravity directly, which is then rendered to the image plane to generate a dense reference depth map. By combining the image and the reference depth, we estimate the depth based on a discriminative neural network, where the residual between the reference depth and the actual depth is explicitly estimated within the network.}
\label{fig_pipeline}
\end{figure}



\begin{figure*}[tpb]
\centering
\includegraphics[height=5.4cm]{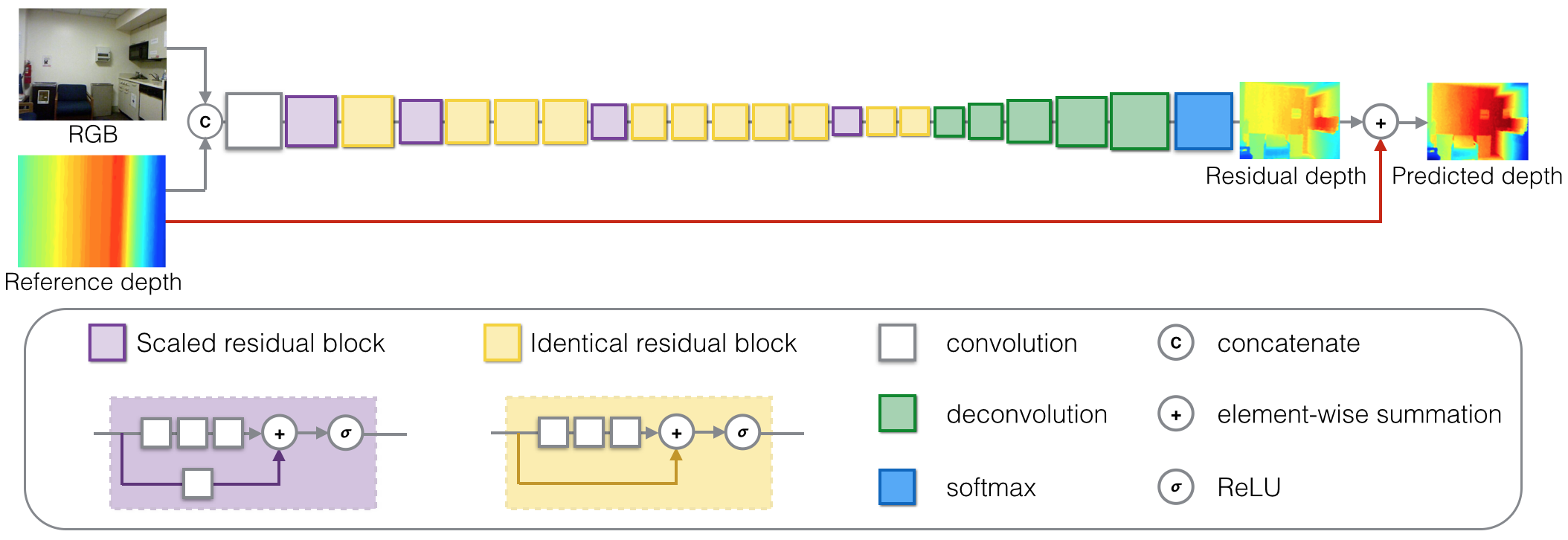}
\caption{ Our residual of residual network architecture. The network is designed based on the ResNet-50, with 50 convolutional layers and 5 deconvolutional layers. We add a global identity skip to send the reference depth to the last feature layer, which is denoted as the red line in the figure. The global identity skip encourages the network to explicitly learn the residual depth. Best viewed in color.}
\label{fig_architecture}
\end{figure*}

\section{Method}\label{sec_model}

In this section,  we present our novel methods for the partial observation task. The pipeline of our method is illustrated in Figure~\ref{fig_pipeline}, which is composed of the input construction and the discriminative neural network.



\subsection{Construction of reference depth}\label{sec_input}  

Given the partially observed depth data with a 2D laser range finder, a naive idea of generating a sparse depth map is by associating the available laser range depths to pixels and rest of all the pixels are padded with zeros~\cite{cadena2016multi}.
Intuitively, the network can hardly learn useful information from the sparse depth map generated using 2D laser range data due to the extremely sparse distribution. On the other hand, such sparse depth map is a mixture of two different categories of values. The zero value filled in the unknown region is the logical code to denote whether there is a valid depth value, while the depth value is a real number for the distance. Mixing the logical code and the depth value in the same map might confuse the network.


To avoid the problems mentioned above, we construct a dense reference map where every pixel is assigned with a depth value, which we name as ``{\textit{reference depth}}'' map. The generation process is visualized in Figure~\ref{fig_pipeline}.
Firstly, median filtering is applied to the laser scan readings for smoothness, followed by interpolation between adjacent laser points. The linear interpolation is employed for imputing the missing depth values.
Secondly, at each point in the imputed laser scan, we generate a line along the gravity direction in 3D space, resulting in a family of lines composing a surface vertical to the ground plane. Finally, by rendering this virtual surface to the image plane of the corresponding monocular camera, we can obtain the dense reference depth map. A related work to our reference depth map is Stixel~\cite{badino2009stixel}, which compactly represents the scene as many vertical rectangular sticks. Their work validates the feasibility of our depth map construction pipeline in real situation. 

After constructing the reference depth map, it is concatenated with the corresponding image as the input of the subsequent neural network for learning. 
Note that with a dense reference depth map as the input, the task of the depth estimator is transformed to {\textit{sculpting a depth value from the reference}}, while it is originally formulated as {\textit{creating a depth value from the unknown}}.
This transformation significantly changes the declaration of the depth estimation problem, which we consider as one crucial reason for boosting the performance.






\subsection{Residual of residual network}\label{sec_residual}

In order to sculpt the depth from the reference map, we want to estimate the difference between the real depth and the reference map, which we formally denote as ``{\textit{residual depth}}''. Here we use the residual neural network (ResNet)~\cite{he2015deep} as our network backbone because of its inherent design to learn the residual, as well as its superior performance on a wide range of computer vision tasks recently. 
It should be noted that, the residual in this paper, is assigned with exact physical concept, which reveals whether the actual depth is closer or further than the reference depth at each pixel. However, in the original ResNet, the network is actually learning a transformed residual since the input experiences nonlinear transformations during the feed-forward propagation. Thus we propose the ``\textit{residual of residual network}'' to explicitly estimate the residual depth, which is particularly suited to the partially observed estimation tasks.

Our network architecture is illustrated in Figure~\ref{fig_architecture}, where the main branch is a fully convolutional network extended from the standard ResNet-50. ResNet is composed of two kinds of residual blocks, i.e. the scaled residual blocks and the identical residual blocks. In both residual blocks, there are two branches, the lower one is an identity skip connection aiming at preserving the information of the block input, and the upper one consists of three convolutional layers which are encouraged to model the residual with respect to the input. Both kinds of blocks can be represented as
\begin{equation}\label{eq_scale}
x_{l+1} = \sigma (F(x_l,W_l) + h(x_l))
\end{equation}
where $x_l$ is the input of the $l$th block,  $W_l$ is the weights of the three convolutional layers of the $l$th block, $\sigma(x) = max(0,x)$ is the nonlinear ReLU layer. For the scaled residual blocks, $h(\cdot)$ is a convolutional layer to scale the feature maps, while in identical residual blocks $h(x)$ is an identity mapping. Considering the nonlinear ReLU layer, as $F(x_l,W_l)$ and $h(x_l)$ are not ensured to be positive, we have the inequality
\begin{equation}\label{eq_relu}
 \sigma (F(x_l,W_l) + h(x_l)) \neq \sigma (F(x_l,W_l)) + \sigma(h(x_l))
\end{equation}
Due to the existence of the nonlinear mapping, the block does not exactly learning the residual $x_{l+1}-x_{l}$ between the input and the output, even though $h(\cdot)$ is the identity mapping in the identical residual block. Consequently, the full network learns a transformed residual between the network input $x_0$ and the final network output $x_L$, rather than the residual depth $x_L-x_0$, where $x_0$ denotes the reference depth and $x_L$ denotes the actual depth in our model. In addition, the deconvolutional layers located between the residual blocks and the network output further transform the feature maps and interrupt our initial attempt for learning the residual depth. 

To encourage the network to explicitly learn to sculpt the depth from the reference depth, we add a global identity skip as the red line in Figure~\ref{fig_architecture} to directly send the reference depth to the last feature map before the output, which can be presented as $x_L= x_{L-1}+x_0$. Thus $x_{L-1}$ is enforced to explicitly learn the residual depth map $x_L-x_0$. Hence there are two categories of identity skips in our network, i.e. the local identity skips connecting each residual block and the global identity skip connecting the full network, that is the reason it is called the residual of residual network.

\subsection{Combination of classification and regression} \label{sec_classregress}

Previous works usually formulate the depth estimation problem as a classification task or a regression task~\cite{eigen2014depth,eigen2015predicting,liu2015learning,laina2016deeper,cao2016estimating}. In this paper, we construct a loss function tightly combining the classification loss and the regression loss. Specifically, a softmax layer is added on top of the final deconvolution layer as visualized in Figure~\ref{fig_architecture}. For classification, all depth values are discretized into $K$ bins, where the center depth value of each bin is denoted as $k$. We set $K=101$ in our experiments. Let us denote the input feature vector of the softmax layer as $f_i$, then the probability of the corresponding sample $i$ being assigned to the discretized depth $k$ is computed as 
\begin{equation}
p_i^k = \frac{exp(f_i^T\theta_k)}{\sum_{k=1}^K{exp(f_i^T\theta_k)}}
\end{equation}
Then the predicted depth is given as
\begin{equation}\label{eq_argmax}
\hat{y}_i = \arg \max_k p_i^k
\end{equation}
As (\ref{eq_argmax}) can only provide a discretized depth estimation, we propose to calculate the predicted depth using the expected value as
\begin{equation}\label{eq_expected}
\bar{y}_i = \sum_{k=1}^{K} p_i^k k 
\end{equation}
By calculating the expected value, we obtain a continuous depth estimation. The expectation is also more robust than the discretized value with the maximal probability. Furthermore, it is also more convenient for calculating the gradients with respect to the expected value (\ref{eq_expected}) compared to the argmax value (\ref{eq_argmax}).

With the predicted probability, the softmax classification loss is given as 
\begin{equation}
L_c =  \sum_{i=1}^M \sum_{k=1}^{K} \delta (\left [y_i \right ]-k) \log(p_i^k)
\end{equation}
where $M$ is the number of all samples, $y_i$ is the ground truth depth and $\left [y_i \right ]$ is the center depth value of the discretized bin that $y_i$ falls in. $\delta(x)=1$ when $x=0$, otherwise $\delta(x)=0$. For regression, we use L1 loss to generate a constant gradient even when the difference is small, which is formulated as 
\begin{equation}
L_r =  \sum_{i=1}^M |y_i - \bar{y}_i|
\end{equation}
Then we combine the classification loss and the regression loss to train the network as
\begin{equation}
L = L_c + \alpha L_r
\end{equation}
where $\alpha$ is the constant weight term. We set $\alpha=1$ in our experiments.

When compared with individual classification or regression losses, our combination of both these two losses brings the following remarks:
\begin{itemize}
\item The classification loss alone cannot distinguish the difference across discretized bins while regression is able to provide larger penalty to the larger predictive errors.
\item If the estimated depth falls into the correct bin, then the classification loss would vanish. The regression loss still works to eliminate the small loss within the bin, leading to a finer estimation.
\item Compared to the solution with regression loss alone, our method can provide a probabilistic distribution. Furthermore, as the depth is computed as the expected value, it has a fixed range and thus is more robust compared to the direct regression.
\end{itemize}



\subsection{Estimation refinement} \label{sec_refine}

As shown in Figure~\ref{fig_architecture}, the network outputs the predicted depth by summing the reference depth and the residual depth, without additional trainable layers. We further refine the predicted depth by applying the median filtering. It can reduce the noises generated from the summation and slightly promote the estimation performance. 

\section{Experiments}\label{sec_exp}
The experiments were preferably carried out on publicly available data sets for benchmarking and easier comparison. In this paper, we evaluate our method on the indoor dataset NYUD2~\cite{silberman2012indoor} and the outdoor dataset KITTI~\cite{Geiger2012CVPR}.

\subsection{Experimental setup}

NYUD2~\cite{silberman2012indoor} is an indoor dataset collected using the Microsoft Kinect. It covers 464 scenes with 4 million raw image and depth pairs. We follow the official split to use 249 scenes for training and 215 scenes for testing. We sampled 50,000 images from the raw training data for training, where the missing depth values were masked out during the training process. Test set includes 654 images with filled-in depth values, which is the same as the other monocular depth estimation methods~\cite{eigen2014depth,eigen2015predicting,liu2015learning,cao2016estimating,laina2016deeper}. We simulated a laser scan that was perpendicular to the gravity direction, with a fixed height above the ground plane. In our experiment, the height was set as 80cm. Since NYUD2 is collected using a hand-held Kinect, the camera pose varies a lot between different frames, leading to an uncertain gravity direction. Thus we follow Gupta. et al~\cite{gupta2013perceptual} to estimate the gravity direction, of which we observe the accuracy is acceptable in our experiments. 

For the outdoor dataset KITTI~\cite{Geiger2012CVPR}, we use three scene categories (``City'', ``Residential'' and ``Road'') in the raw data for training and testing, the same as Eigen et al.~\cite{eigen2014depth}. We sampled 5,000 images captured by the left color camera from 30 scenes for training, and 632 images from other 29 scenes for testing. With the relative small training set, the network is initialized with the weights learned from NYUD2. The ground truth depth was obtained with a Velodyne HDL-64E 3D laser scanner, where the missing depth was masked out for both training and testing. As the Velodyne laser scanner observed 64 laser scans in each frame, we simulated a 2D laser range finder by taking one of the laser scans as our partially observed data. The laser scan was selected to be within a fixed range of polar angle in the spherical coordinate, which was set as $88 ^{\circ} \pm 2 ^{\circ}$ in our experiment. As the sensors were fixed on a mobile car in the KITTI, the gravity direction was fixed for all frames and could be obtained from the offline calibration. It is to be noted that the gravity direction is also fixed in practical applications of both indoor and outdoor robots, and there is no requirement for the additional estimation of the gravity direction.

For the network configuration, both image and reference depth in NYUD2 were reshaped as $320\times256$, and the predicted size was $160\times128$.  The input size of KITTI was set as $320\times96$, with output size $160\times48$. Though the depth is only available at the half bottom of the image in KITTI, we input the full image into the network for learning the context. Note that the predicted result was up-scaled to the original size for evaluation on both NYUD2 and KITTI.

We implemented our residual of residual network based on Caffe~\cite{jia2014caffe}. Following ResNet, we also used batch normalization for efficient convergence and the batchsize was set as 16. The loss was summed over all valid pixels and the learning rate $\eta=10^{-6} \times 0.98^{\left \lfloor n/1000 \right \rfloor}$, $n$ denotes the iteration number. The model was trained for 80,000 iterations, which took about 33 hours on a Nvidia Titan X. Following~\cite{eigen2014depth}, we used online data augmentation to avoid over-fitting. Specifically, the data augmentation steps include rotation, scaling, color transformation and flips. 

We used the following standard metrics to evaluate our performance, where $y_i$ is the ground truth depth, $\bar{y}_i$ is the estimated depth value and $N$ is the number of total pixels:
\begin{itemize}
\begin{spacing}{1.3}
\item Root Mean Squared Error (rms): $ \sqrt{\frac{1}{N} \sum_i (\bar{y}_i-y_i)^2}$
\item Mean Absolute Relative Error (rel): $\frac{1}{N} \sum_i \frac{|\bar{y}_i-y_i|}{y_i}$
\item Mean log10 Error (log10): $\frac{1}{N} \sum_i |\log_{10}\bar{y}_i-\log_{10}y_i|$
\item Threshold $\delta_k$: percentage of $y_i$, s.t. $\max(\frac{\bar{y}_i}{y_i}, \frac{y_i}{\bar{y}_i}) < \delta^k$, $\delta=1.25$ and $k=1,2,3$.
\end{spacing}
\end{itemize}

\subsection{Model evaluation}
\begin{table*}[h]
\caption{Model evaluation on NYUD2.}
\label{table_valid}
\begin{center}
\begin{tabularx}{.9\textwidth}{@{}lclc|YYY|YYY@{}}

\hline
\multirow{2}{*}{Input}  &   \multirow{2}{*}{Res. of Res.}    & \multirow{2}{*}{Loss}   & \multirow{2}{*}{Refined} &\multicolumn{3}{c|}{Error (\textit{lower is better})}     &    \multicolumn{3}{c}{Accuracy (\textit{higher is better})} \\
                        &                            &   &  & rms & rel & log10     &$\delta_1$&$\delta_2$&$\delta_3$\\
\hline
RGB                     &   --                     & C.      & -- &0.642  & 0.184 	&0.071	& 76.2 &92.7 &97.4 \\  
RGB                     &   --                     & C.+R.   & -- &0.617  & 0.173	&0.068  & 77.2 &93.8 &97.8 \\  
RGB + Ref.           	&   No                     & C.   	 & -- &0.537  & 0.124   &0.051  & 86.2 &95.1 &97.9 \\
RGB + Ref.           	&   No                     & C.+R.   & -- &0.507  & 0.126   &0.050  & 86.3 &95.7 &98.4 \\   
RGB + Ref.           	&   Yes                    & C.      & No &0.480  & 0.108   &0.045  & 87.0 &95.8 &98.5 \\
RGB + Ref.           	&   Yes                    & C.+R.   & No &0.451  & 0.106   &0.044  & 87.4 &96.2 &98.8 \\
RGB + Ref.           	&   Yes                    & C.+R.   & Yes&0.442  & 0.104   &0.043  & 87.8 &96.4 &98.9 \\
\hline
\end{tabularx}
\end{center}
\end{table*}

\begin{table*}[h]
\caption{Comparison with the state-of-the-art on NYUD2 and KITTI.}
\label{table_nyud2}
\begin{center}
\begin{tabularx}{.8\textwidth}{@{}l|l|YYY|YYY@{}}

\hline
\multirow{2}{*}{Dataset} & \multirow{2}{*}{Method}    &  \multicolumn{3}{c|}{Error (\textit{lower is better})}     &    \multicolumn{3}{c}{Accuracy (\textit{higher is better})} \\
                       &   & rms & rel & log10       & $\delta_1$&$\delta_2$&$\delta_3$\\
\hline

\multirow{6}{*}{NYUD2} & Liu et al.~\cite{liu2015learning} & 0.824 & 0.230 & 0.095& 61.4& 88.3& 97.1\\
&Eigen et al.~\cite{eigen2014depth} 		&0.907	&0.215  &--		&61.1 &88.7 &97.1 \\
&Eigen et al.~\cite{eigen2015predicting}	&0.641 	&0.158	&--		&76.9 &95.0 &98.8\\
&Cao et al.~\cite{cao2016estimating}		&0.645 	&0.150	&0.065	&79.1 &95.2 &98.6\\
&Laina et al.~\cite{laina2016deeper}		&0.583	&0.129	&0.056	&80.1 &95.0 &98.6\\
&Ours            							&\textbf{0.442} &\textbf{0.104}	&\textbf{0.043}	&\textbf{87.8} &\textbf{96.4} &\textbf{98.9}  \\  
\hline
\multirow{4}{*}{KITTI} & Saxena et al.~\cite{saxena2009make3d} 	& 8.734&  	0.280&	--&		60.1&	82.0&	92.6\\
&Eigen et al. ~\cite{eigen2014depth}		& 7.156& 	0.190&	--&		69.2&	89.9&	96.7\\
&Mancini et al.~\cite{mancini2016fast}   	& 7.508&	--&		--&		31.8&	61.7&	81.3\\
&Cadena et al.~\cite{cadena2016multi}		& 6.960&	0.251&	--& 	61.0&	83.8&	93.0 \\	
&Ours 										& \textbf{4.500}&	\textbf{0.113}&	\textbf{0.049}&		\textbf{87.4}&	\textbf{96.0}&	\textbf{98.4}\\
\hline
\end{tabularx}
\end{center}
\end{table*}
To demonstrate the benefits gained from the reference map construction, the network design and the refinement process, we conducted comparison experiments on NYUD2 against some variants of our proposed method. Table~\ref{table_valid} illustrates the comparison results. Note that all variants listed in the table were implemented based on ResNet-50 with the same network capacity. Specifically, we first performed the monocular depth estimation using only RGB image as a baseline. Then the laser information was added to the input as the reference depth map (``Ref.''), without the residual of residual structure (``Res. of Res.''). Furthermore, we added the global identity skip to explicitly estimate the residual depth. Finally, the refinement was performed to refine the predicted depth of our residual of residual network. All network architectures were trained in two different loss function settings, the classification loss alone (``C'') or the combination of classification and regression (``C.+R.''). As can be seen from the table, \\
\begin{itemize}
\item By comparing the results in the first two rows with the followings, it can be seen that the performance is substantially promoted with our reference depth map as the additional input. It validates the effectiveness of constructing a dense reference map from the sparse partial observation  as described in Section~\ref{sec_input}, which redefines the depth estimation task as sculpting the depth from the reference. 
\item Comparison between the results with and without the “Res. of Res.” structure demonstrates the superiority by adding the global identity skip. As we explained in Section~\ref{sec_residual}, the ResNet is inherently suited to our partially observed task as we want to learn the residual depth, adding the global identity skip further preserves the exact physical concept of the residual, resulting in the boosts of the estimation performance. 
\item By comparing the performances with classification loss alone and with combination of classification and regression, it can be seen that the combination raises the performances in all model variants. It demonstrates the benefit of our loss design as introduced in Section~\ref{sec_classregress}.
\item The refinement introduced in Section~\ref{sec_refine} further brings slight improvement to the estimation accuracy.
\end{itemize}

\begin{figure*}[tpb]
\centering
\subfigure{\includegraphics[height=1.5cm]{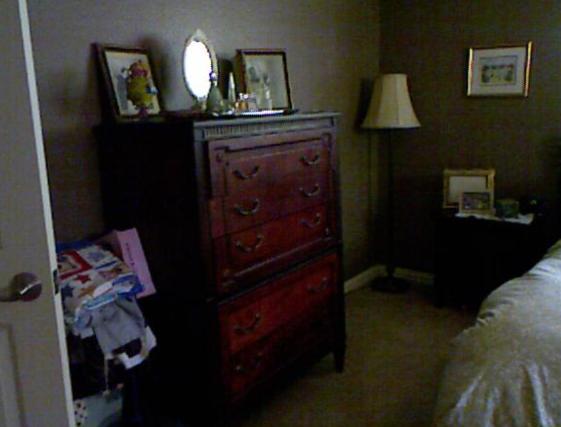}}
\subfigure{\includegraphics[height=1.5cm]{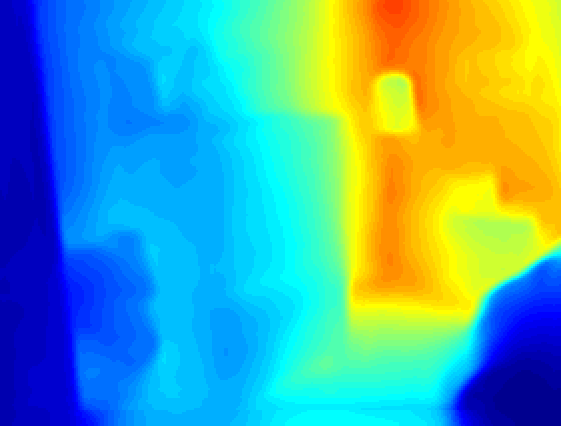}}
\subfigure{\includegraphics[height=1.5cm]{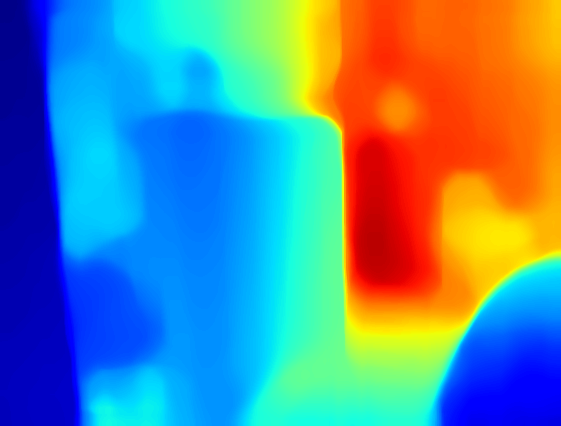}}
\subfigure{\includegraphics[height=1.5cm]{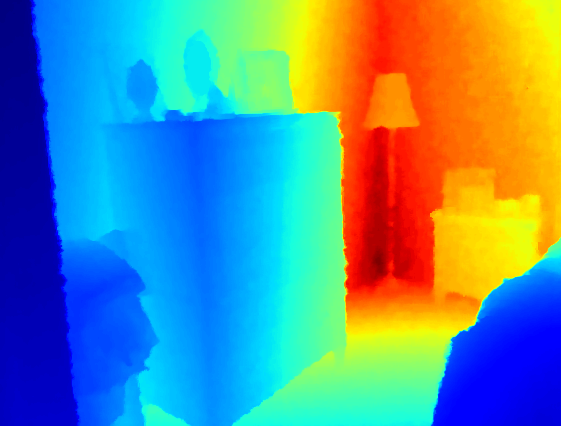}}
\subfigure{\includegraphics[height=1.5cm]{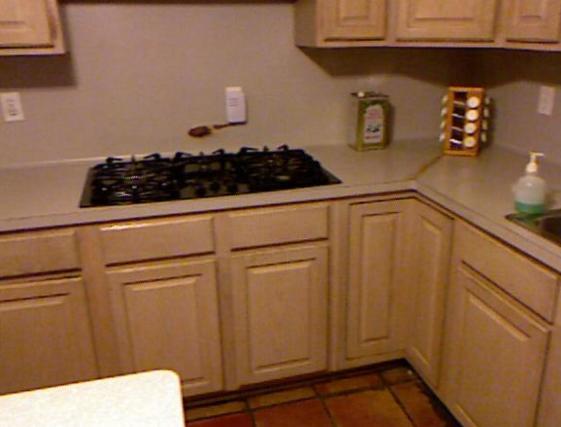}}
\subfigure{\includegraphics[height=1.5cm]{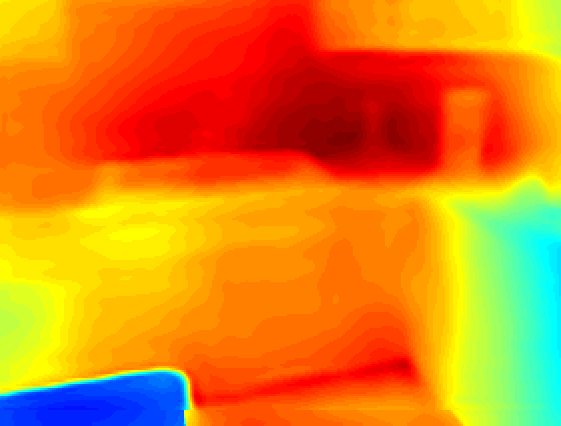}}
\subfigure{\includegraphics[height=1.5cm]{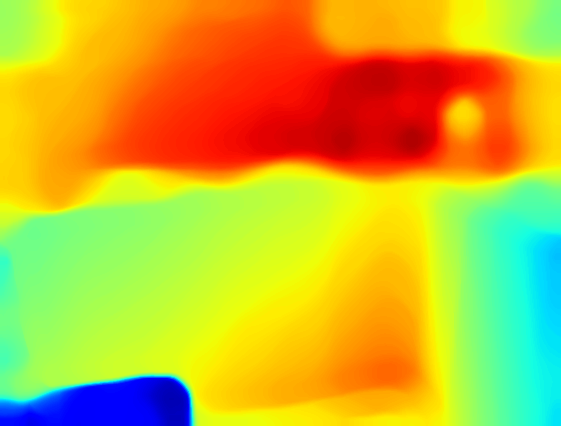}}
\subfigure{\includegraphics[height=1.5cm]{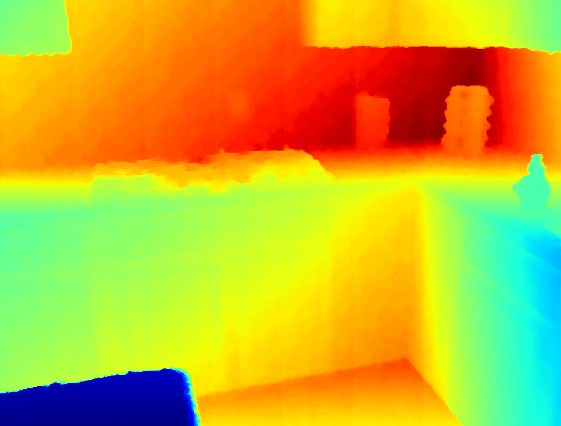}}
\subfigure{\includegraphics[height=1.5cm]{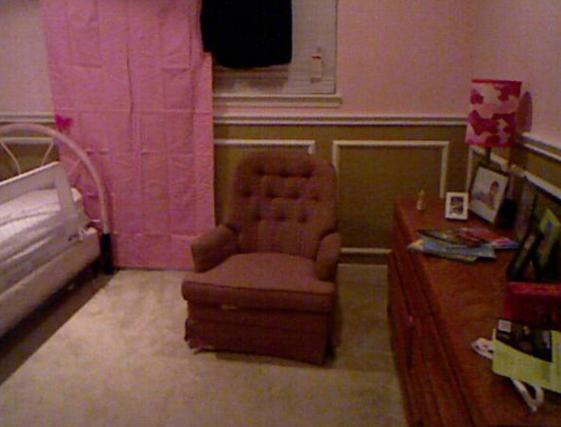}}
\subfigure{\includegraphics[height=1.5cm]{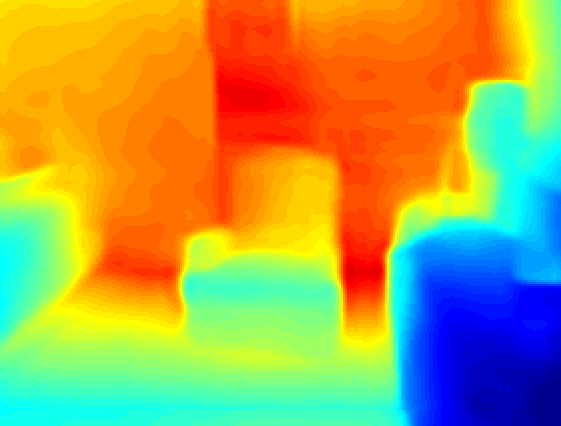}}
\subfigure{\includegraphics[height=1.5cm]{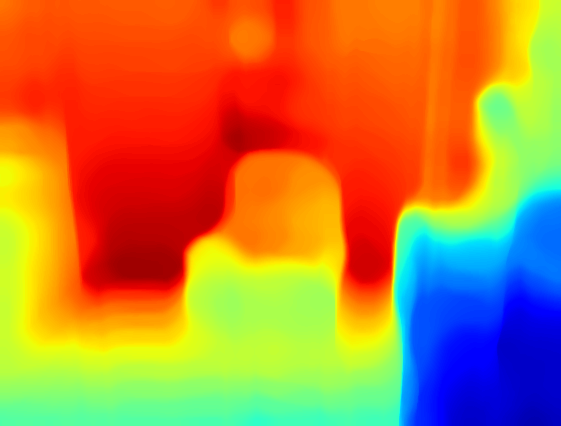}}
\subfigure{\includegraphics[height=1.5cm]{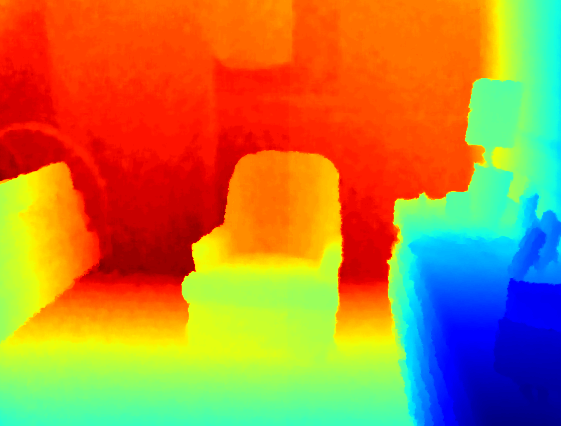}}
\subfigure{\includegraphics[height=1.5cm]{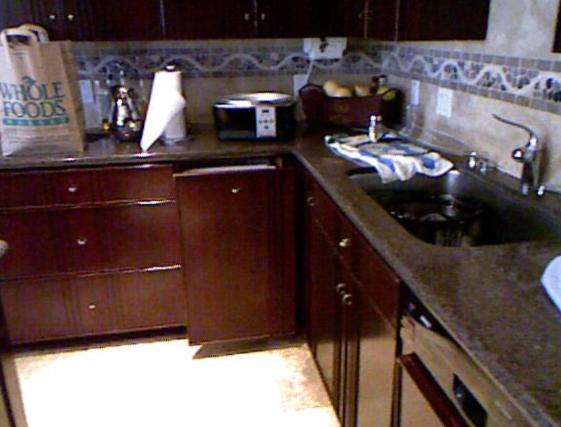}}
\subfigure{\includegraphics[height=1.5cm]{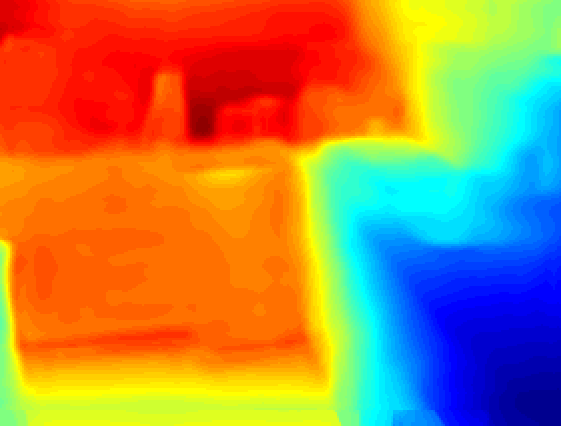}}
\subfigure{\includegraphics[height=1.5cm]{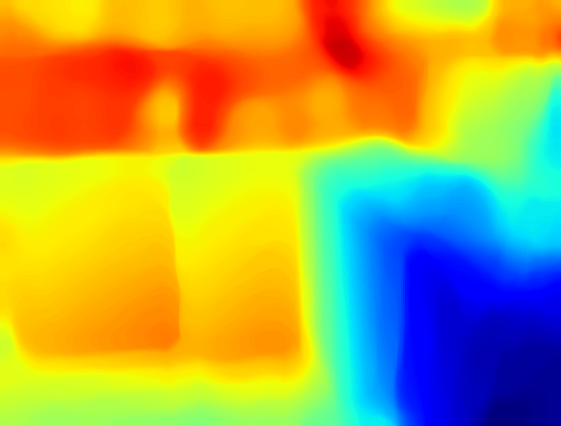}}
\subfigure{\includegraphics[height=1.5cm]{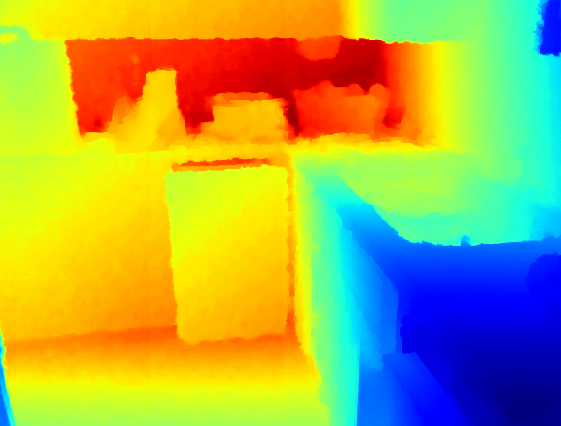}}
\stackunder[1pt]{\includegraphics[height=1.5cm]{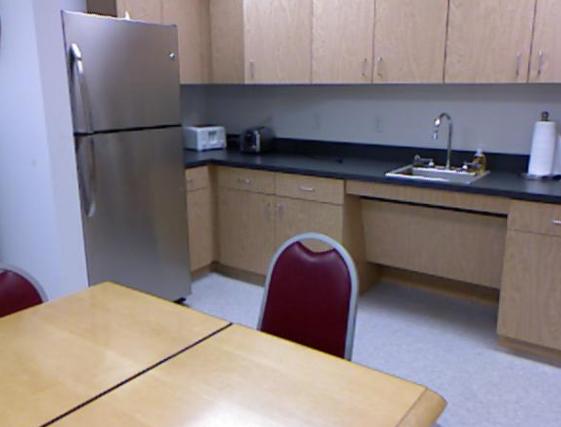}}{\small{Image}}
\stackunder[1pt]{\includegraphics[height=1.5cm]{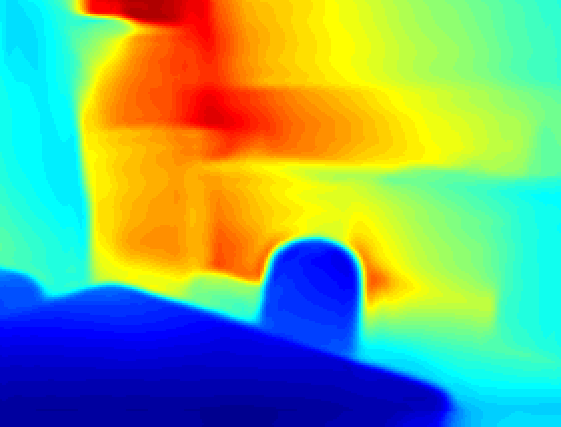}}{\small{Eigen et al.}}
\stackunder[1pt]{\includegraphics[height=1.5cm]{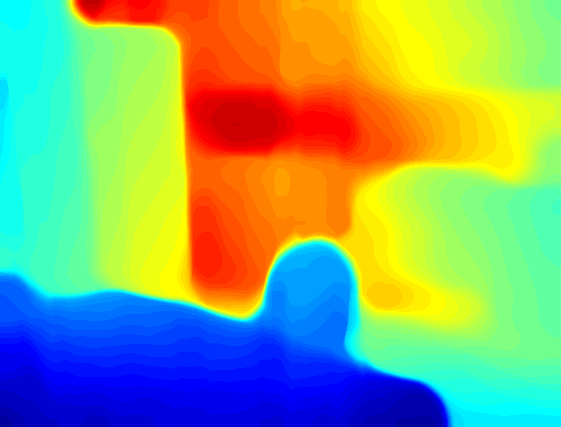}}{\small{Ours}}
\stackunder[1pt]{\includegraphics[height=1.5cm]{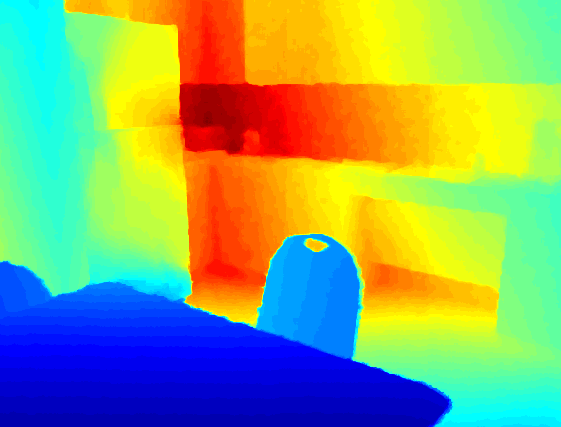}}{\small{Ground truth}}
\stackunder[1pt]{\includegraphics[height=1.5cm]{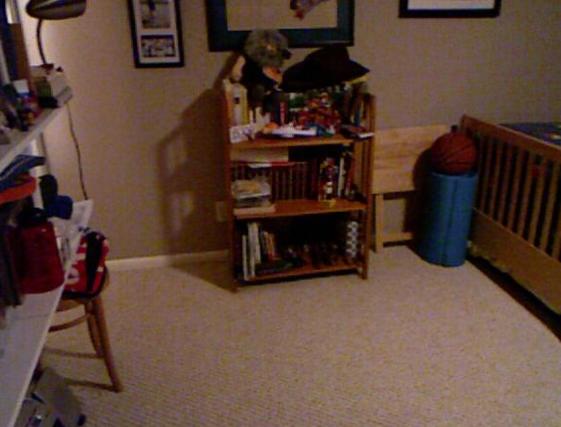}}{\small{Image}}
\stackunder[1pt]{\includegraphics[height=1.5cm]{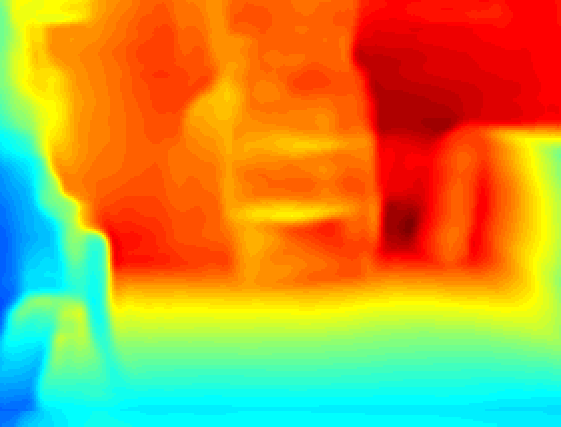}}{\small{Eigen et al.}}
\stackunder[1pt]{\includegraphics[height=1.5cm]{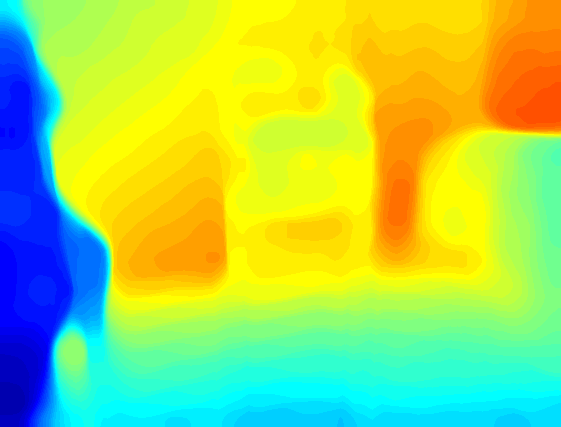}}{\small{Ours}}
\stackunder[1pt]{\includegraphics[height=1.5cm]{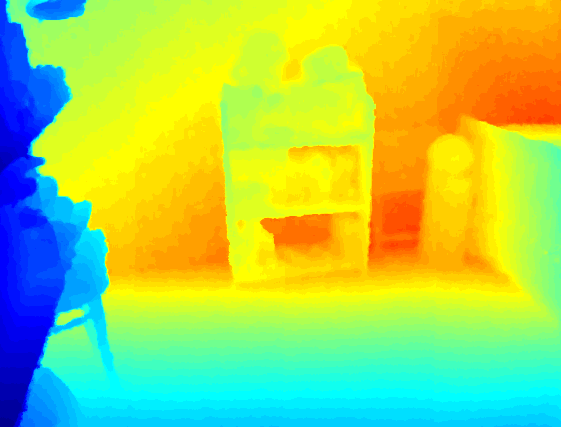}}{\small{Ground truth}}
\caption{Comparison on NYUD2. For each test image, the corresponding depth estimated by Eigen et al.~\cite{eigen2015predicting}, our method and the ground truth are given from left to right. Red denotes far and blue denotes close in the depth maps. It can be seen that the estimation of Eigen et al.~\cite{eigen2015predicting} usually has a bias due to the ambiguity of the global scale.
Best viewed in color. }
\label{fig_samples_nyu}
\end{figure*}

\begin{figure*}[tpb]
\centering
\subfigure{\includegraphics[height=0.92cm]{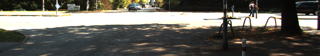}}
\subfigure{\includegraphics[height=0.92cm]{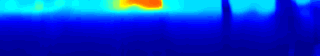}}
\subfigure{\includegraphics[height=0.92cm]{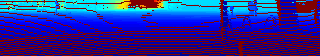}}
\subfigure{\includegraphics[height=0.92cm]{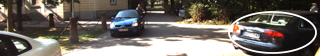}}
\subfigure{\includegraphics[height=0.92cm]{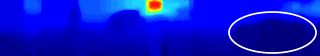}}
\subfigure{\includegraphics[height=0.92cm]{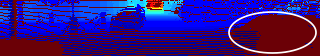}}
\subfigure{\includegraphics[height=0.92cm]{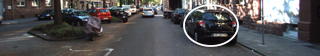}}
\subfigure{\includegraphics[height=0.92cm]{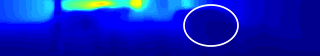}}
\subfigure{\includegraphics[height=0.92cm]{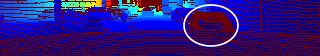}}
\stackunder[1pt]{\includegraphics[height=0.92cm]{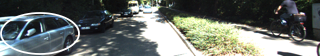}}{\small{Image}}
\stackunder[1pt]{\includegraphics[height=0.92cm]{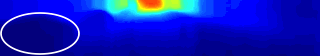}}{\small{Ours}}
\stackunder[1pt]{\includegraphics[height=0.92cm]{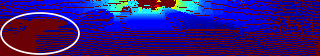}}{\small{Ground truth}}

\caption{Comparison on KITTI. For each test image, the corresponding depth estimated our method and the ground truth are given from left to right. Red denotes far and blue denotes close in the depth images, and missing values are denoted as dark red. The white circles demonstrate that missing depth value usually occurs due to the reflection when using the 3D laser range finder, while our method gives a reliable depth estimation. Best viewed in color.}
\label{fig_samples_kitti}
\end{figure*}

\subsection{Comparison with the state-of-the-art}
In Table~\ref{table_nyud2}, we compared with the state-of-the-art depth estimation methods using our best model suggested in Table~\ref{table_valid}. We conducted experiments on both NYUD2 and KITTI to validate the generalization ability of the proposed method. Quantitative results in Table~\ref{table_nyud2} illustrates that the depth estimation accuracy is substantially promoted by adding a single planar view of laser range data, validating our hypothesis of resolving the scale ambiguity with laser sensor information. It is to be noted that Cadena et al.~\cite{cadena2016multi} also tackled the depth estimation task with partially available depth. They proposed to reconstruct the dense depth from a sparse depth map with the unknown depth padded with a constant, while we formulate the partially observed estimation task as a residual learning task by constructing a dense reference map and achieves a considerable promotion.

Corresponding qualitative comparisons are given in Figure~\ref{fig_samples_nyu} and Figure~\ref{fig_samples_kitti}. For NYUD2, we compared with Eigen et al.~\cite{eigen2015predicting} and the ground truth in Figure~\ref{fig_samples_nyu}. As can be seen, taking the advantage of a single planar of laser data, our methods parse the scenes better with a more accurate estimation of the global scale. Figure~\ref{fig_samples_kitti} illustrates the comparison between our method and the ground truth obtained from the Velodyne laser range finder. Images and depth maps are cropped to show the region with laser observations. It is to be noted that the 3D laser range finder usually cannot observe valid depth values when scanning the windows of the cars, which might lead to unsafe predictions in the high-level decision tasks. On the contrary, our method provides a relatively stable depth estimation regardless of the reflection.

\subsection{Analysis and potential usage}

\begin{figure*}[tpb]
\centering
\subfigure[rms]{\includegraphics[height=2.3cm]{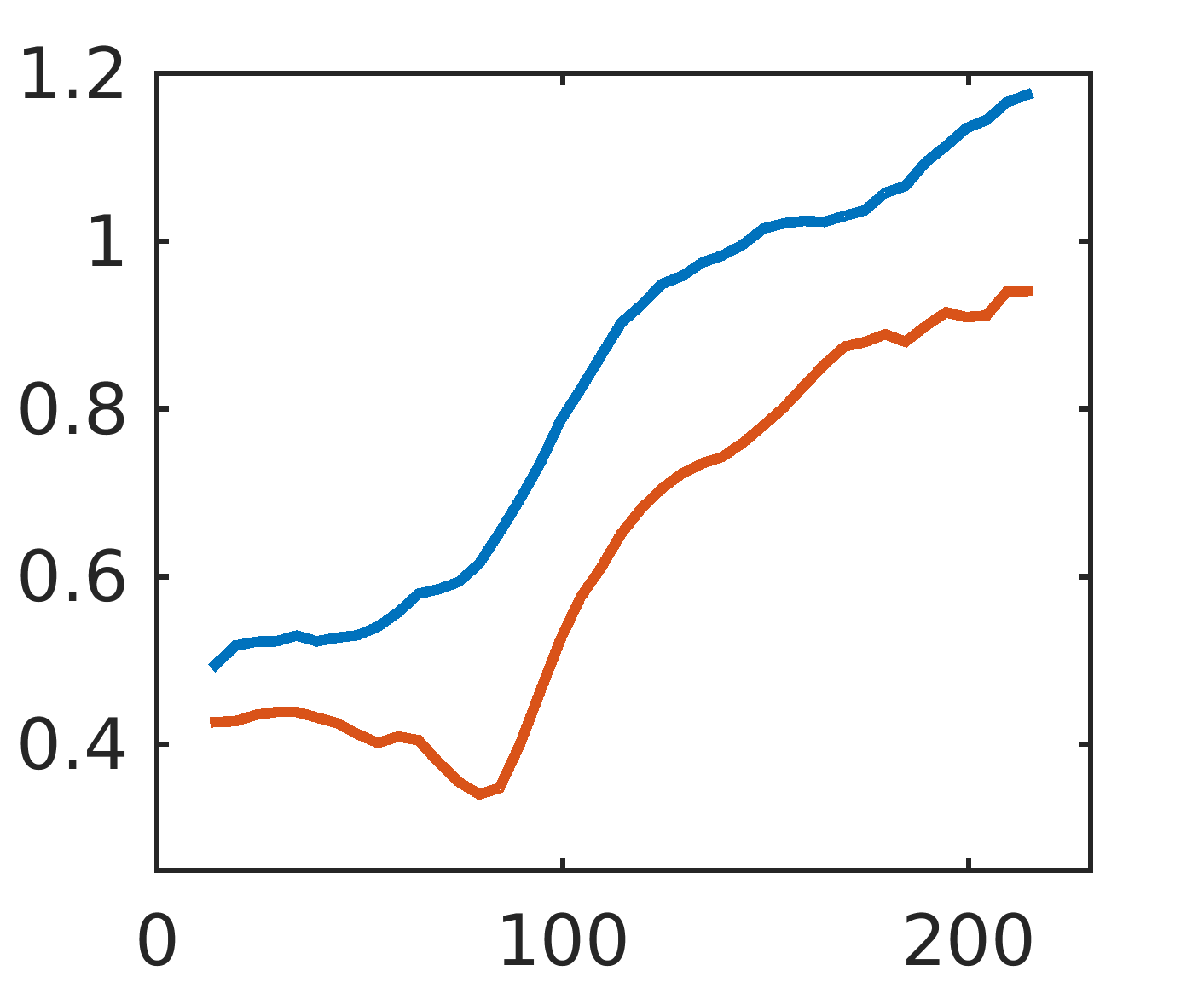}\label{fig6_rms}}
\subfigure[rel]{\includegraphics[height=2.3cm]{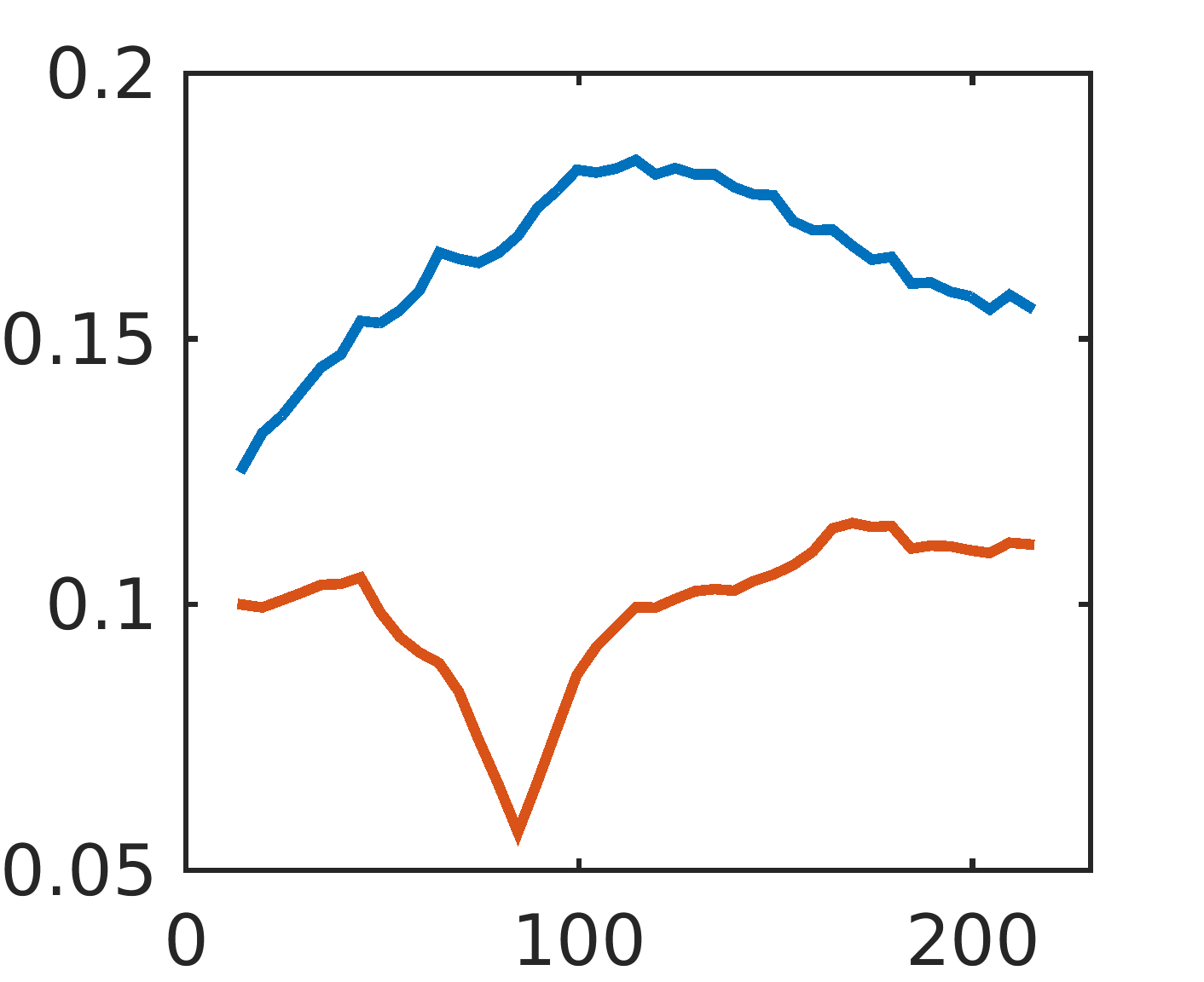}\label{fig6_rel}}
\subfigure[log10]{\includegraphics[height=2.3cm]{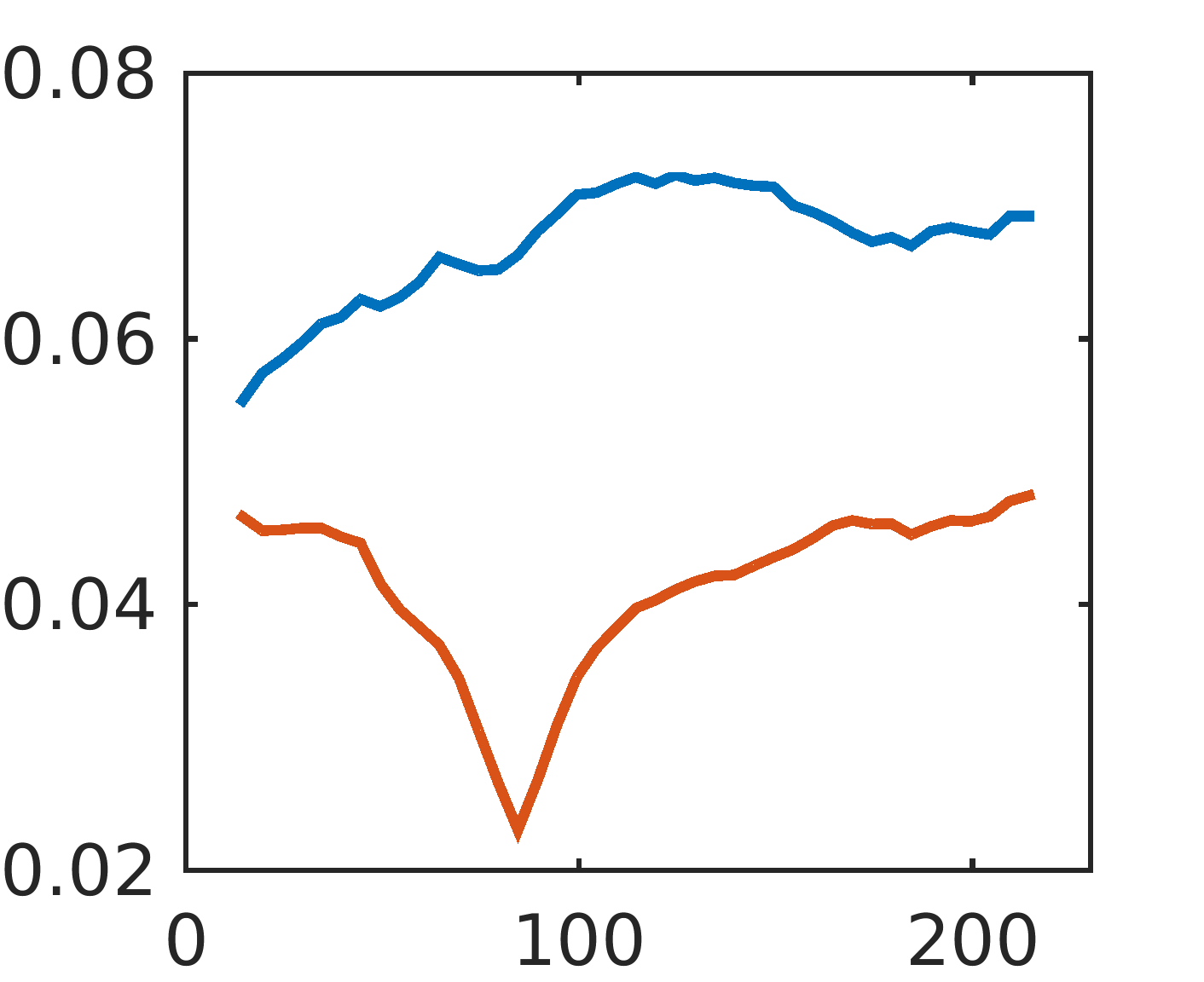}\label{fig6_log}}
\subfigure[$\delta_1$]{\includegraphics[height=2.3cm]{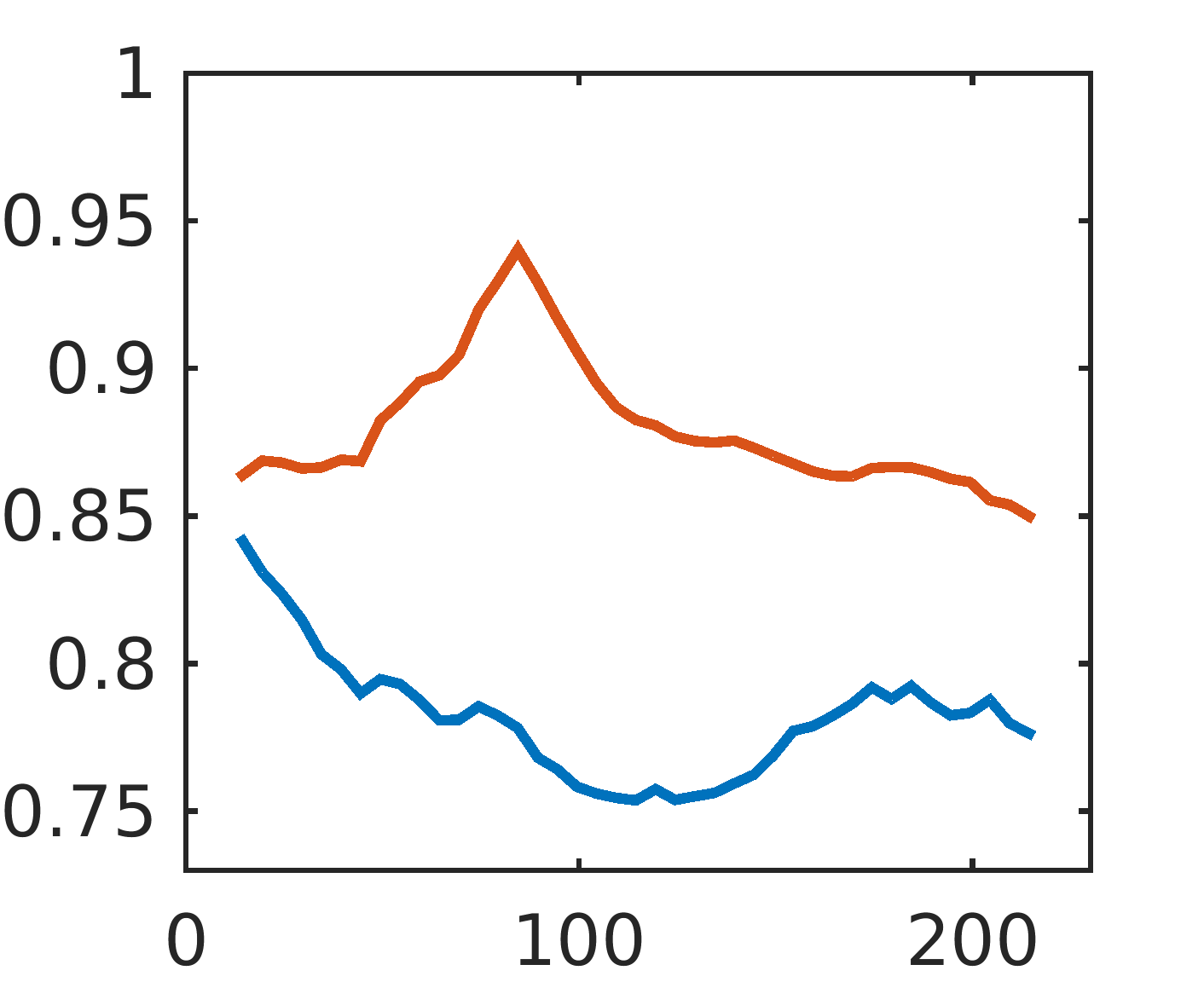}\label{fig6_awt1}}
\subfigure[$\delta_2$]{\includegraphics[height=2.3cm]{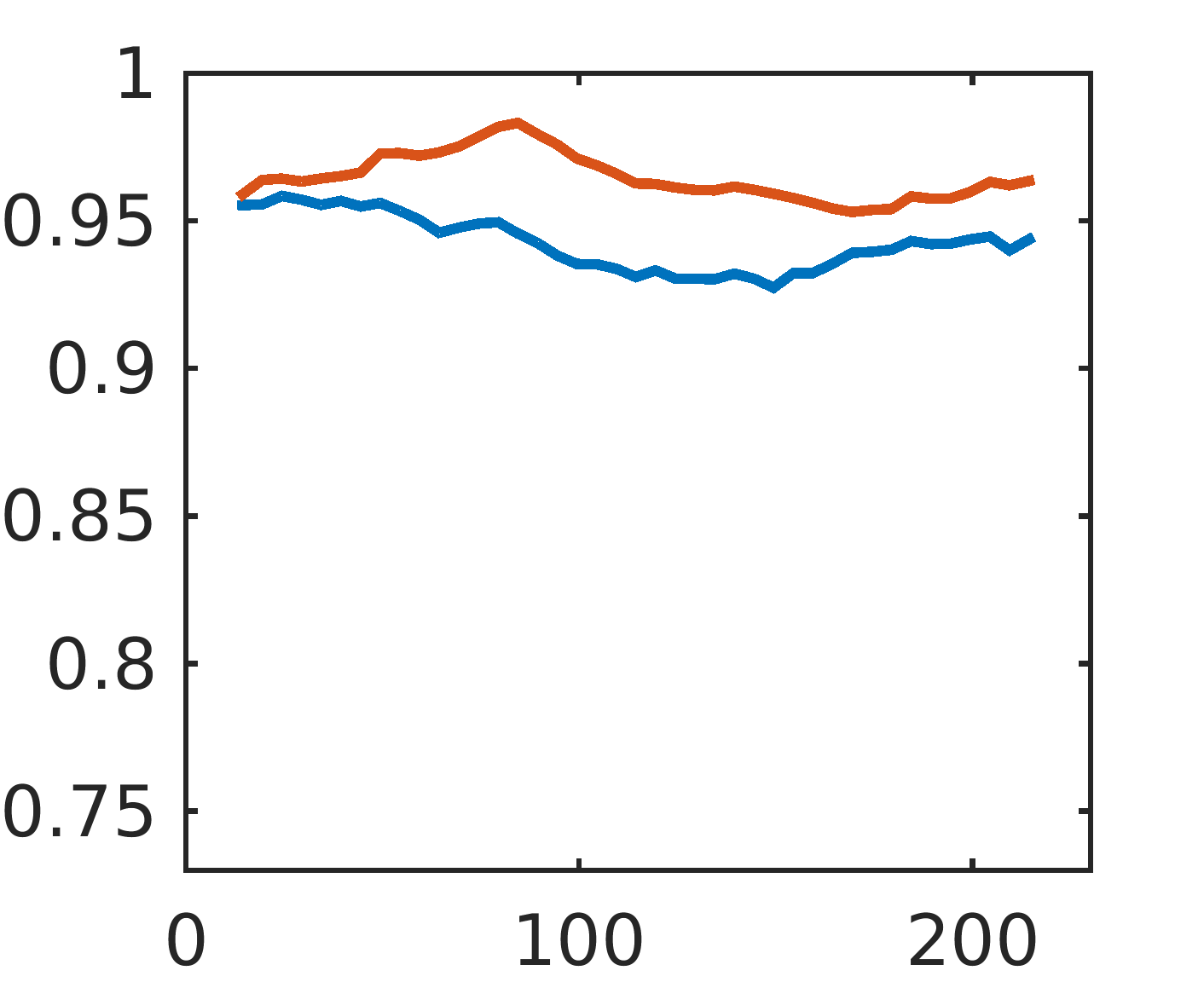}\label{fig6_awt2}}
\subfigure[$\delta_3$]{\includegraphics[height=2.3cm]{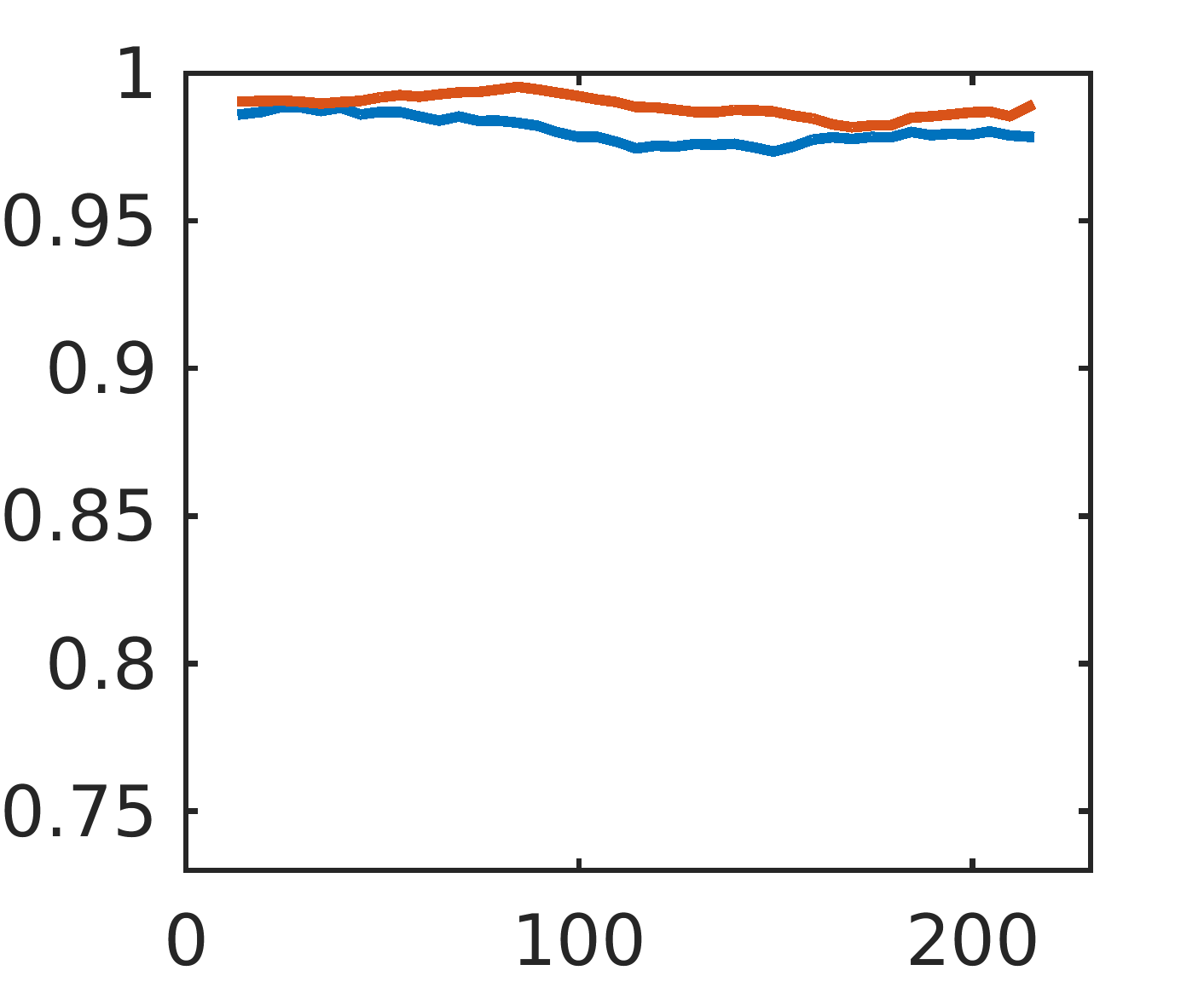}\label{fig6_awt3}}
\caption{Evaluation results at different heights. In each figure, the x-axis is the height in centimeter, and the y-axis is the evaluation metric. The blue line denotes the performance of the depth estimation with only RGB image, and the red line denotes the performance of the depth estimation with combination of the RGB image and a single line of laser range scan with height at 80cm. Best viewed in color.}
\label{fig_rgblasercompare}
\end{figure*}

To reveal the intrinsic impact of introducing the limited and sparse partial observation, we analyzed the depth estimation performances at different heights of the scene on NYUD2. Specifically, we generated a set of scans that are perpendicular to the gravity direction and sampled above the ground from 10cm to 210cm at equal distances. For all test images on NYUD2, the same evaluation metrics mentioned above were applied to evaluate on those individual scans.
Figure~\ref{fig_rgblasercompare} reveals the comparison between our monocular depth estimation results (second row in Table~\ref{table_valid}) and our refined partially observed result (last row in Table~\ref{table_valid}). It is to be noted that there is a minimum point in each error metric in Figure~\ref{fig6_rms},~\ref{fig6_rel},~\ref{fig6_log}, and a corresponding maximum point in each accuracy metric in Figure~\ref{fig6_awt1},~\ref{fig6_awt2},~\ref{fig6_awt3}. This is corresponding to the height of our laser scan (80cm), which demonstrates that the observed laser information is effectively preserved based on our residual of residual neural network.
Furthermore, the partial observation not only increases the performance at the height of the 2D laser range finder, but also gives a considerable promotion to the performances of the overall scene. In addition, Figure~\ref{fig_rgblasercompare} also indicates the possibility of further promotion by adding more 2D laser range scans to the other heights.


\begin{figure*}[tpb]
\centering
\subfigure{\includegraphics[height=2.65cm]{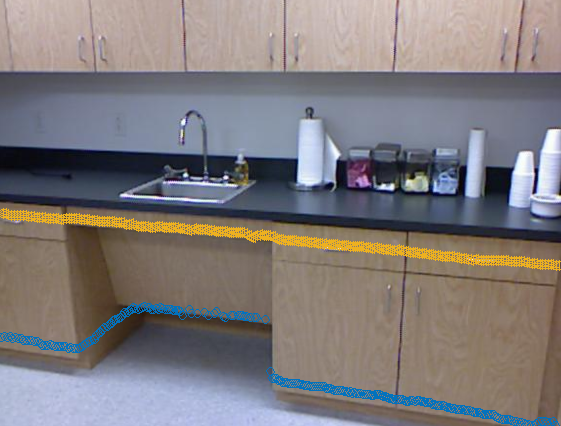}}
\subfigure{\includegraphics[height=2.65cm]{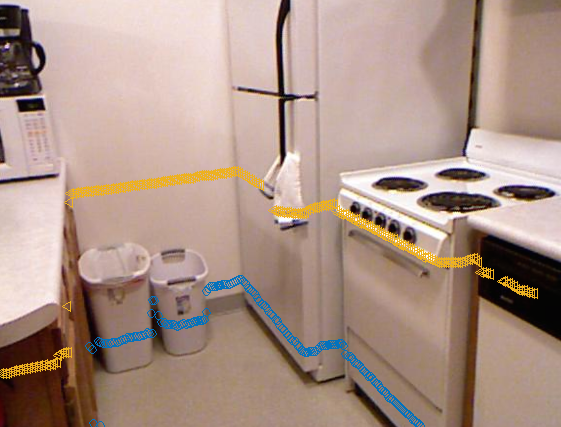}}
\subfigure{\includegraphics[height=2.65cm]{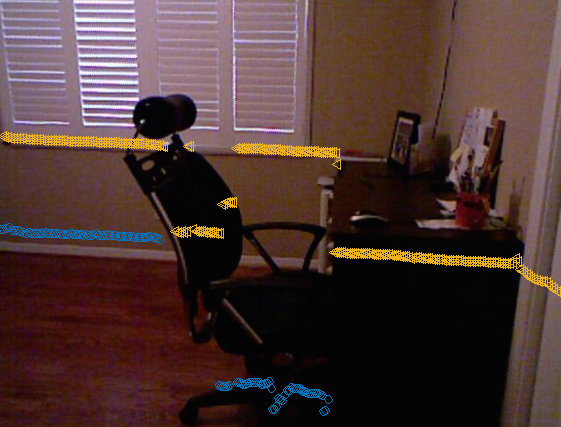}}
\subfigure{\includegraphics[height=2.65cm]{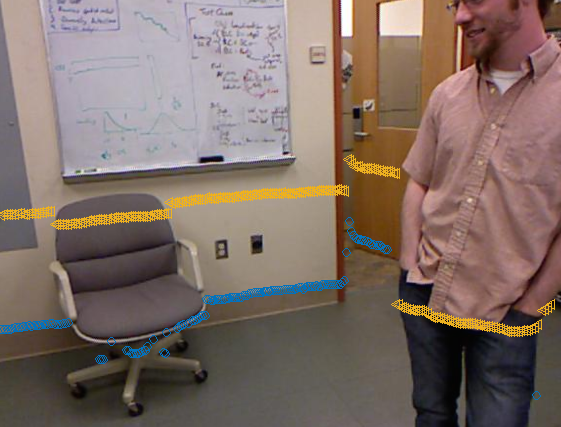}}
\subfigure{\includegraphics[height=2cm]{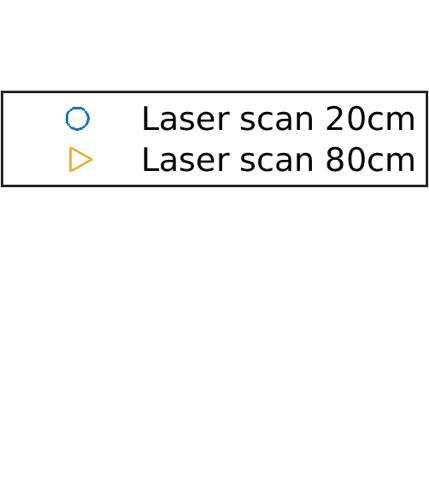}}
\stackunder[0.5pt]{\includegraphics[height=2.87cm]{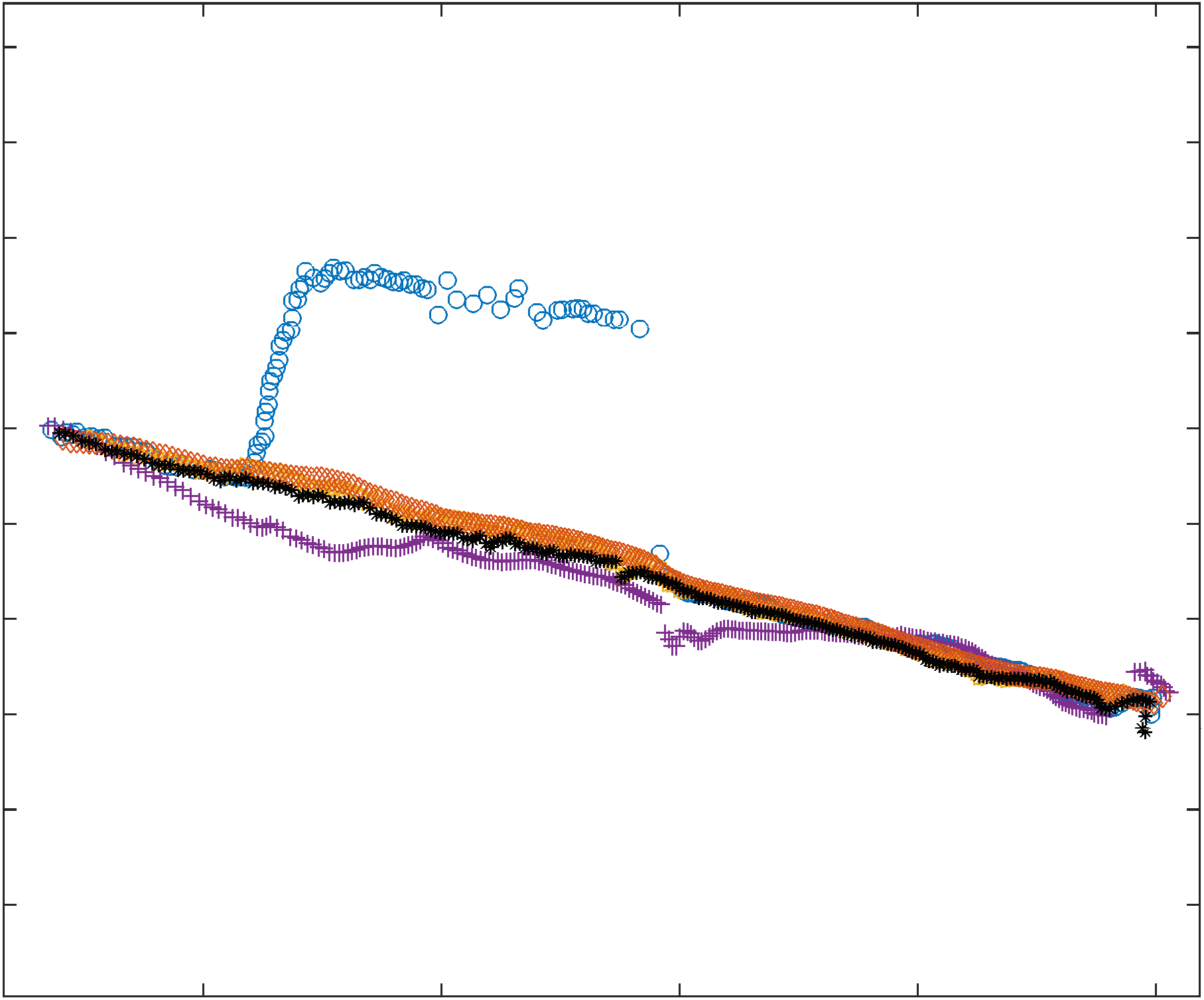}}{\footnotesize{(a)}}
\stackunder[0.5pt]{\includegraphics[height=2.87cm]{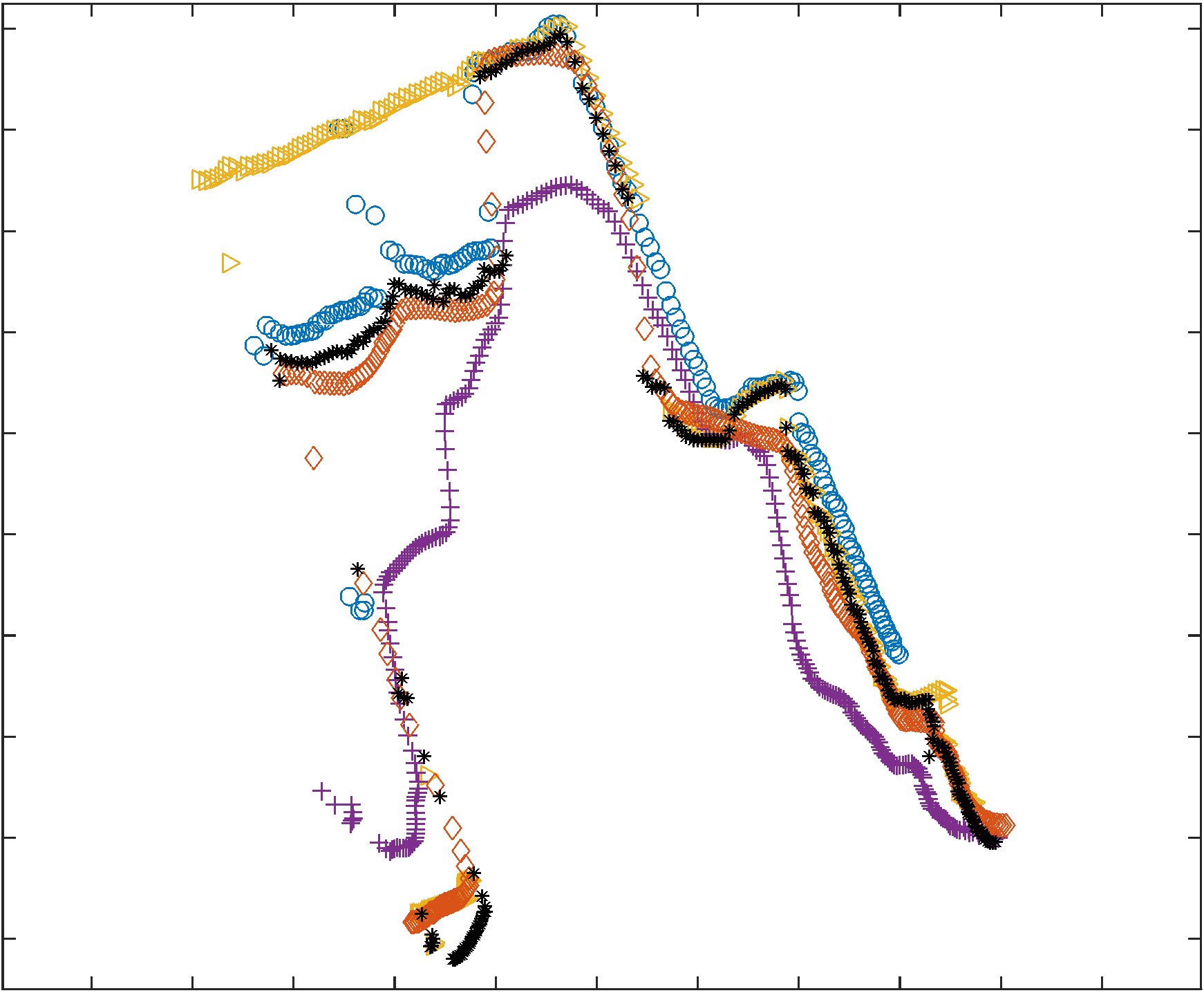}}{\footnotesize{(b)}}
\stackunder[0.5pt]{\includegraphics[height=2.87cm]{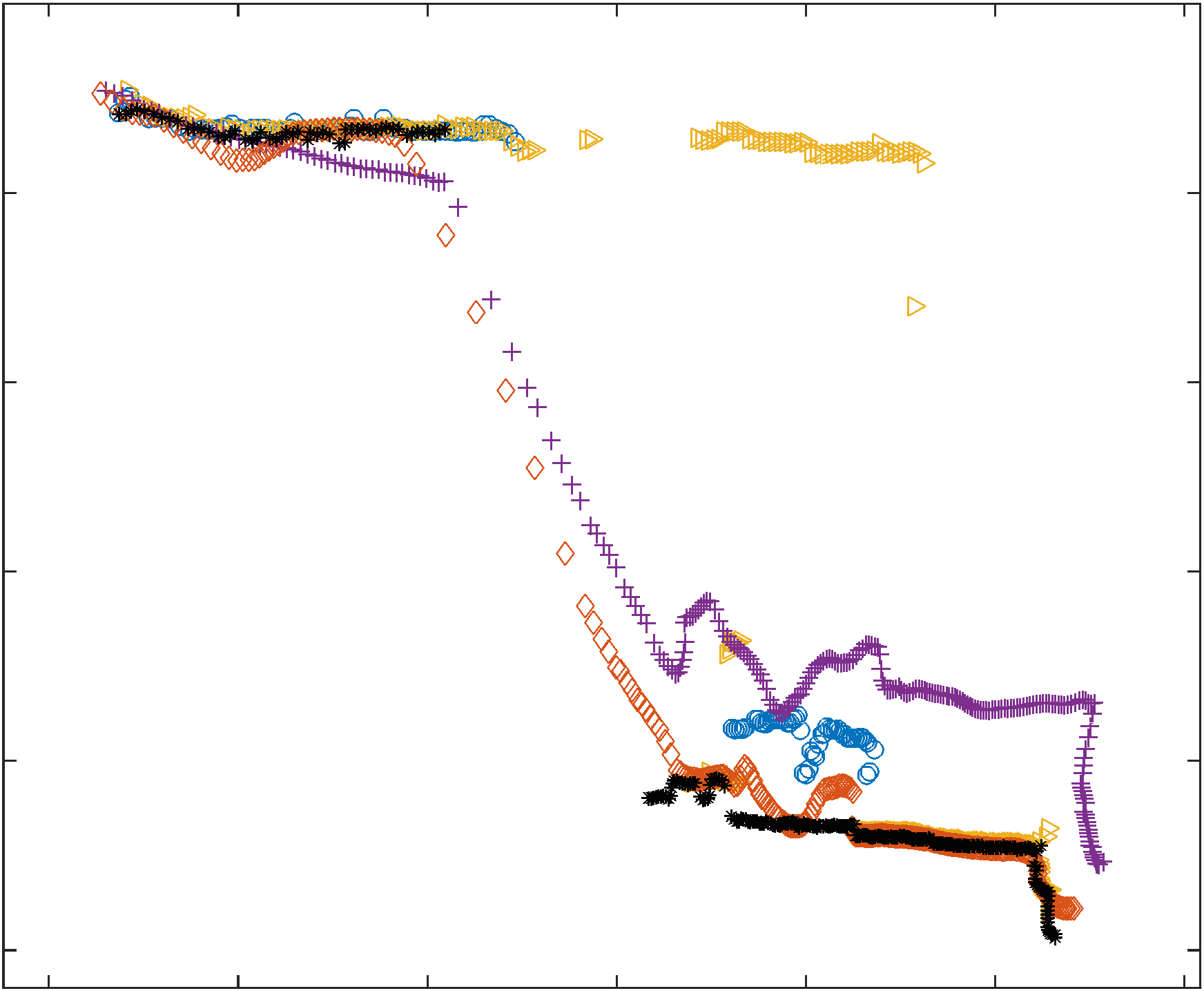}}{\footnotesize{(c)}}
\stackunder[0.5pt]{\includegraphics[height=2.87cm]{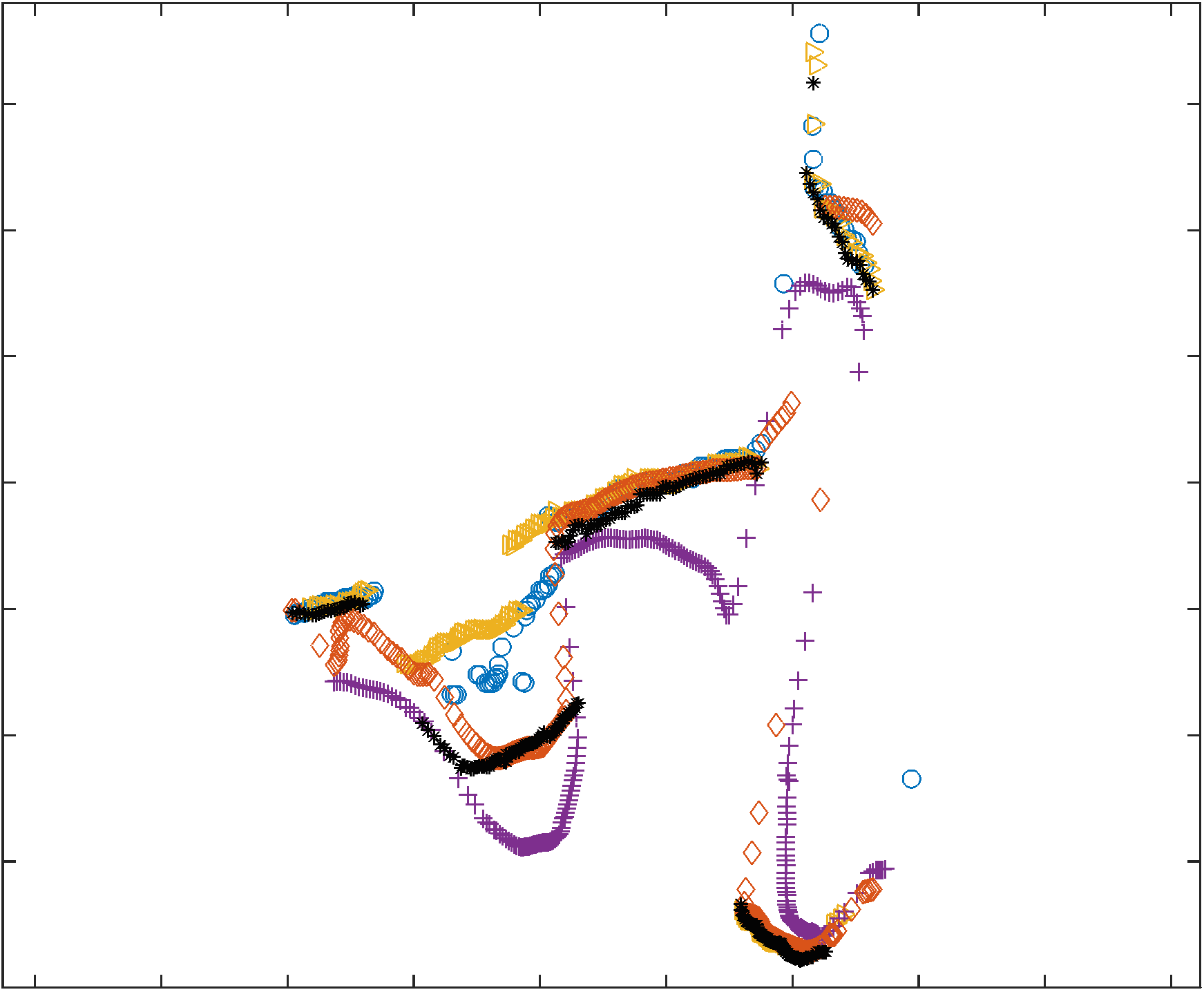}}{\footnotesize{(d)}}
\subfigure{\includegraphics[height=2.15cm]{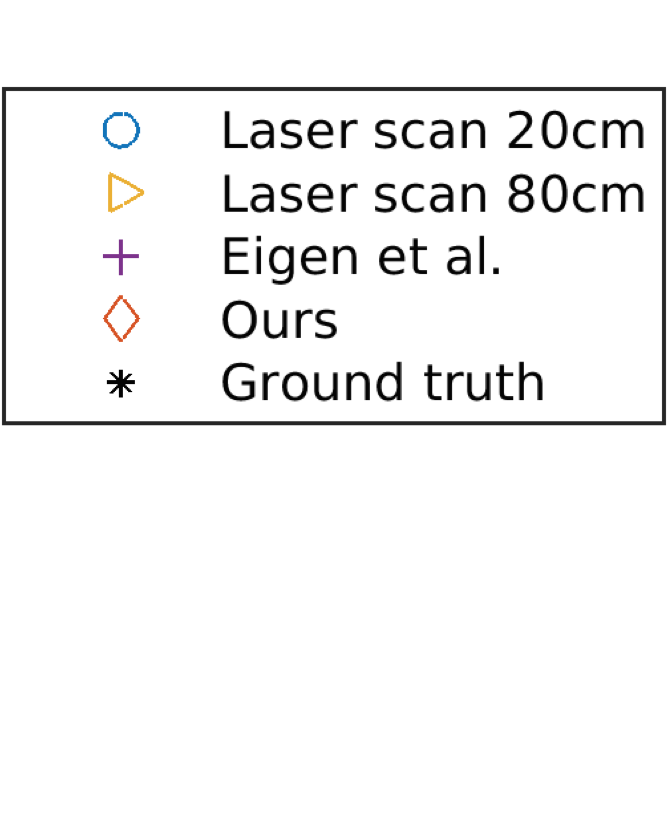}}
\caption{ Illustration of obstacle avoidance. The first row shows the image with the projected laser scan. The height of the laser is set at 20cm to  simulate the real situation. The second row demonstrates the corresponding obstacle maps in the two-dimensional space, with the gravity direction eliminated. Best viewed in color.}
\label{fig_obstacle}
\end{figure*}

Figure~\ref{fig_rgblasercompare} also demonstrates our method has a reliable root mean squared error below the heights of 80cm. This is particularly suited to the robotics applications. We demonstrate that our method has the potential for obstacle avoidance in Figure~\ref{fig_obstacle}. 
Specifically, we parsed the geometry based on different methods for comparison. This includes the simulated 2D laser range finder, the estimated dense depth and the ground truth depth provided by the Kinect. Following the general method for obstacle avoidance based on point clouds, we projected each dense depth map to the 3D space, and then down-projected all the 3D points within the height $(0,M]$ to the 2D plane to obtain the nearest obstacle in the scene. Here 0 is the height of the ground and $M$ is a safe range that is usually set to be higher than the robot. We set $M=100$cm in this example. The simulated laser range scan was also set to be perpendicular to the gravity direction, which can be directly presented in the 2D plane. For thorough comparison, we simulated two laser range finders at 20cm and 80cm above the ground plane respectively. 
Figure~\ref{fig_obstacle} demonstrates the images and the corresponding obstacle maps generated from different methods. As can be seen, the laser scanner set at 20cm fails to detect the upper stove and the seat of the chairs in Figure 7(a), 7(c) and 7(d), while the laser scanner set at 80cm misses the 
lower garbage bins as well as the seats in Figure 7(b), 7(c) and 7(d). These failure detections might lead to collision with the obstacles in practical applications, which suggests that it is challenging for reliable obstacle avoidance using a fixed 2D laser range finder due to its limited view, indicating the importance of understanding the 3D geometry. For depth estimation with only monocular images, there is usually a bias caused by the scale ambiguity. Our method relieve these problems by estimating the dense depth based on the 2D laser data, leading to a comprehensive depth estimation with higher reliability for obstacle avoidance.

\section{Conclusion}\label{sec_conclu}
This paper explores the monocular depth estimation task. By introducing sparse 2D laser range data into the depth estimation task, our method effectively alleviates the global scale ambiguity and produces a more reliable estimation result. We redefined the depth estimation task as a residual learning problem by constructing a dense reference map from the sparse laser range data. It was implemented with our residual of residual network, and the classification and regression losses are combined for more effective estimation. We conducted experiments on both indoor and outdoor datasets including NYUD2 and KITTI. The performance was compared with state-of-the-art techniques and our method shows superior results on both datasets, validating the effectiveness of the proposed method. Furthermore, it suggests a promising direction to use sparse laser data to guide dense depth estimation using learning methods.

\bibliographystyle{ieeetr}
\bibliography{root}

\end{document}